\documentclass[lettersize,journal]{IEEEtran}
\usepackage{amsmath,amsfonts}
\usepackage{algorithmic}
\usepackage{algorithm}
\usepackage{array}
\usepackage[caption=false,font=normalsize,labelfont=sf,textfont=sf]{subfig}
\usepackage{textcomp}
\usepackage{stfloats}
\usepackage{color}
\usepackage{url}
\usepackage{verbatim}
\usepackage{graphicx}
\usepackage{cite}

\newcommand{\jh}[1]{{\color{black}#1}}

\newcommand{\zz}[1]{{\color{red}#1}}

\usepackage{subcaption}
\usepackage{booktabs}

\begin{document}

\title{SkelGen4D: Weakly-Supervised Skeleton-Based 4D Generation for Text-Driven Mesh Animation}

\author{Hao Feng*, Zhi Zuo*, Jia-Hui Pan, Ka-Hei Hui, Zhengzhe Liu$^\dagger$, \\
Dian Zhang, Haoran Xie, \textit{Senior Member, IEEE}, Bin Sheng, and Jingyu Hu
\thanks{*Hao Feng and Zhi Zuo contributed equally to this work.}
\thanks{This work has been supported by National Natural Science Foundation of China (Category C) fund code 62506149 , Lingnan University StartUp Grant fund code: SUG-001/2526, and Faculty Research Grant fund code:106106 and 106119. \emph{(Corresponding author: Zhengzhe Liu$\dagger$.)}}
\thanks{Hao Feng, Zhi Zuo, Zhengzhe Liu, Dian Zhang, and Haoran Xie are with the School of Data Science, Lingnan University, Hong Kong SAR, China.}
\thanks{Jiahui Pan, and Jingyu Hu are with the Department of Computer Science and Engineering, Chinese University of Hong Kong, Hong Kong SAR, China. }
\thanks{Ka-Hei Hui is with Autodesk Research, Canada.}
\thanks{Bin Sheng is with Institute of Computer Applications, Shanghai Jiao Tong University, Shanghai, China.}
}





\maketitle
\begin{figure*}
    \centering
    \includegraphics[width=1\linewidth]{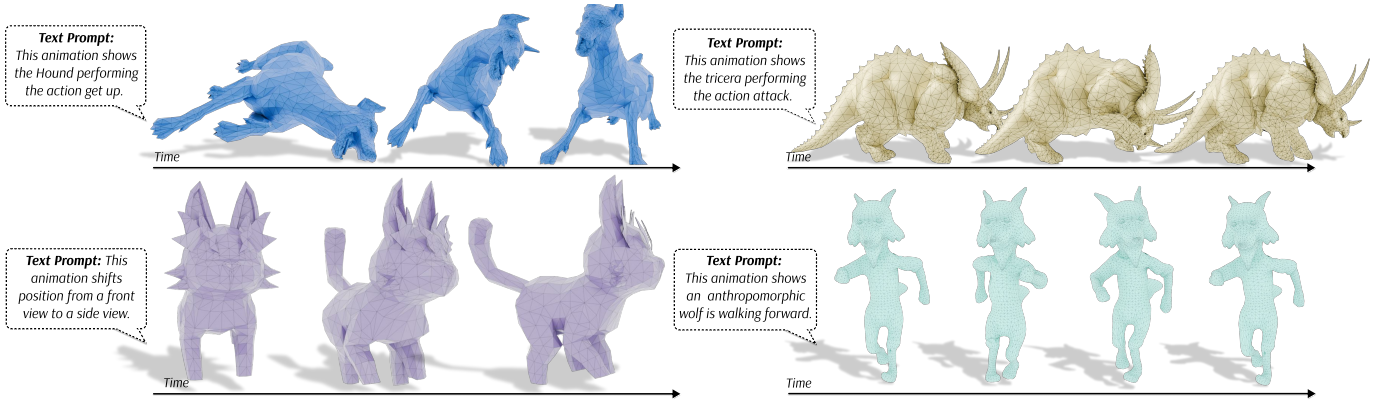}
\caption{
Text-driven mesh animation results generated by \textbf{SkelGen4D} across diverse object categories and motion patterns.
}
     \label{fig:teaser}
\end{figure*}
\begin{abstract}
We study 4D generation to synthesize temporally coherent sequences of 3D geometry for animation and content creation.
In contrast to existing SDS-based optimization methods and video-driven animation approaches, we adopt a skeleton-driven animation framework aligned with standard industrial pipelines, which enables explicit control and editing.
To this end, we propose \textbf{SkelGen4D}, a weakly supervised feed-forward framework for text-driven mesh animation that generates explicit skeleton motions without requiring per-frame skeleton annotations.
SkelGen4D first recovers temporally consistent pseudo-skeletons from animated meshes via differentiable fitting, and then generates text-conditioned skeleton motion sequences in a feed-forward manner, further refined with Motion-GRPO to ensure temporally coherent, physically plausible, and articulated animation.
We evaluate our method on two large-scale benchmarks, Truebones Zoo and Diffusion4D. 
Our results show that our weakly supervised skeleton modeling matches or surpasses fully supervised baselines while scaling to diverse object categories for high-quality text-driven mesh animation. Further, our method supports flexible motion editing and is aligned with standard animation production pipelines.
\end{abstract}


\maketitle

\section{Introduction}

4D generation aims to synthesize temporally coherent 3D geometry sequences and is increasingly important for animation and digital content creation.
Beyond visual realism, practical animation must satisfy temporal coherence, kinematic plausibility, and editability, which substantially increases the difficulty compared to static 3D generation.
In modern industrial production pipelines, these requirements are addressed through skeleton-driven animation, where a hierarchical skeleton rig controls mesh deformation via skinning and binding, enabling precise motion editing, recombination, and integration with downstream tools in film and game engines~\cite{unreal_skeletal_mesh_pipeline, skeletal_animation_guide}.

Recent works on 4D generation mainly differ in how motion is represented and controlled. One line of work distills diffusion priors into dynamic 3D representations via per-instance optimization~\cite{ren2023dreamgaussian4d,ling2024align}. While capable of producing plausible results, these methods are computationally expensive and difficult to scale.
Another line of work animates 3D content by directly modeling mesh deformation, either from video observations or from text-conditioned generative models~\cite{chen2025v2m4,wu2025animateanymesh,song2025puppeteer,huang2025animax,gong2025mocapanything,li2025akd}. Although these approaches improve efficiency and can capture complex dynamics, motion is typically represented implicitly in deformation fields or mesh trajectories, which makes editing and precise control more difficult in standard animation workflows.
In contrast, skeleton-based methods explicitly model articulated motion using structured kinematic representations. Representative works such as AnyTop~\cite{gat2025anytop} adopt a production-style skeleton representation, but require per-frame skeleton motion annotations, which limits scalability. These limitations motivate us to develop SkelGen4D, a skeleton-driven 4D generation framework that aligns with standard production pipelines while avoiding the need for per-frame skeleton annotations.

Our SkelGen4D is a two-stage framework. 
The first stage converts raw mesh animations into temporally consistent pseudo-skeleton trajectories, and the second stage learns a text-conditioned skeleton motion generator from these pseudo-skeletons.
Specifically, in the first stage, starting from a single-frame automatic rigging~\cite{zhang2025unirig}, we derive temporally consistent skeleton motion from raw mesh animation sequences.
\jh{
Instead of applying automatic rigging independently to each frame, which would create temporally inconsistent skeleton topologies, we maintain a consistent skeleton topology.
We retain the initial rig's skinning weights and propagate the skeleton across time by optimizing per-frame joint transformations using gradient descent, ensuring that the skinned mesh deformation matches the observed mesh geometry at each frame.
}
%
This process converts raw mesh dynamics into temporally aligned pseudo-skeleton trajectories without requiring per-frame skeleton annotations.

The second stage is a feed-forward skeleton motion generation framework. Specifically, we first encode motions into a compact latent space and then train a transformer-based generator to produce full sequences conditioned on text, enabling efficient and controllable animation without per-asset optimization.
To further improve temporal consistency beyond the frame-level training described above, we introduce Motion-GRPO, an animation-aware preference objective, inspired by recent advances in large language models~\cite{shao2024deepseekmath}. \jh{Our Motion-GRPO enforces smoothness, motion diversity, and bone-length consistency.}

We evaluate our method on the Truebones Zoo dataset and the Diffusion4D dataset.
On Truebones Zoo, our approach achieves performance comparable to or better than fully supervised methods that rely on per-frame skeleton annotations, in terms of animation fidelity and diversity. 
On Diffusion4D, which covers a wide range of object categories and articulation patterns, our method outperforms video-driven animation baselines and consistently produces high-quality mesh animations with stable motions, demonstrating strong generalization across diverse categories.
Moreover, our explicit skeleton-based representation enables flexible motion editing and aligns well with standard film and game production pipelines.

Overall, we make the following contributions:
\begin{itemize}
    \item We propose a weakly supervised framework for mesh animation that learns skeleton motion directly from raw mesh sequences, without per-frame skeleton annotations.
    \item We develop a feed-forward, text-conditioned skeleton motion generator with an animation-aware Motion-GRPO objective, producing coherent skeleton-controlled animations.
    \item Extensive experiments on Truebones Zoo and Diffusion4D \jh{show that our method achieves} high-quality animation, surpassing video-driven methods \jh{and} competitive with fully supervised skeleton-based approaches, with flexible motion controllability and industrial pipeline compatibility.
\end{itemize}


\section{Related Work}

Recent works on 4D generation mainly differ in how motion is represented and controlled. 

\paragraph{Optimization-based 4D Generation.}
Several works model dynamic 3D content using implicit representations optimized with diffusion priors via Score Distillation Sampling (SDS)~\cite{ren2023dreamgaussian4d,ling2024align,bahmani2024fourdfy,jiang2024consistent4d,zeng2024stag4d}. These approaches optimize dynamic NeRFs or Gaussian primitives on a per-instance basis. While capable of producing high-quality results, they require expensive optimization and suffer from limited scalability.

\paragraph{Deformation-based 4D Generation and Animation.}
Another line of work directly models geometry deformation to animate 3D content, either from video observations or text-conditioned generative models~\cite{sun2024eg4d,yang2024diffusion2,xie2024sv4d,yao2025sv4d2,wu2024cat4d,zhang2024fourdiffusion,ren2024l4gm,jiang2024animate3d,uzolas2025motiondreamer,chen2025v2m4,wu2025animateanymesh,li2025ss4d,song2025puppeteer,huang2025animax,gong2025mocapanything,li2025akd,ActionMesh2026,gong2025swit4dslidingwindowtransformerlossless}. 
These methods typically represent motion implicitly via deformation fields or directly manipulate mesh vertices. While they improve efficiency and can capture complex dynamics, the implicit nature of motion representation makes fine-grained control, local editing, and motion composition more difficult, limiting their compatibility with standard animation pipelines.

\paragraph{Skeleton-based 4D Generation.}
Skeleton-based approaches explicitly model articulated motion using structured kinematic representations. Early works mainly focus on human motion generation, including MotionCLIP~\cite{tevet2022motionclip}, MDM~\cite{tevet2022mdm}, MotionDiffuse~\cite{zhang2022motiondiffuse}, MotionGPT~\cite{jiang2023motiongpt}, and LMM~\cite{zhang2024lmm}, as well as parametric human models such as SMPL~\cite{loper2015smpl}. However, these approaches rely on human-specific priors and fixed skeletal templates, limiting their generalization to diverse articulated objects.

More recent works, such as AnyTop~\cite{gat2025anytop}, extend skeleton-based generation to arbitrary topologies, but require dense per-frame skeleton motion annotations, which limits scalability. These limitations motivate us to develop SkelGen4D, a skeleton-driven 4D generation framework that aligns with standard production pipelines while avoiding the need for per-frame skeleton annotations.

\paragraph{Skeleton Generation and Rigging}
Skeleton generation aims to derive explicit articulation structures for animating 3D models.
Going beyond early approaches~\cite{au2008contraction,huang2013medial,xu2022morig},
recent learning-based methods leverage large-scale 3D data and autoregressive modeling to enable scalable, template-free rigging across diverse categories~\cite{song2025magicarticulate,liu2025riganything,zhang2025unirig}.

\section{Methodology}
\begin{figure*}
    \centering
    \includegraphics[width=0.95\linewidth]{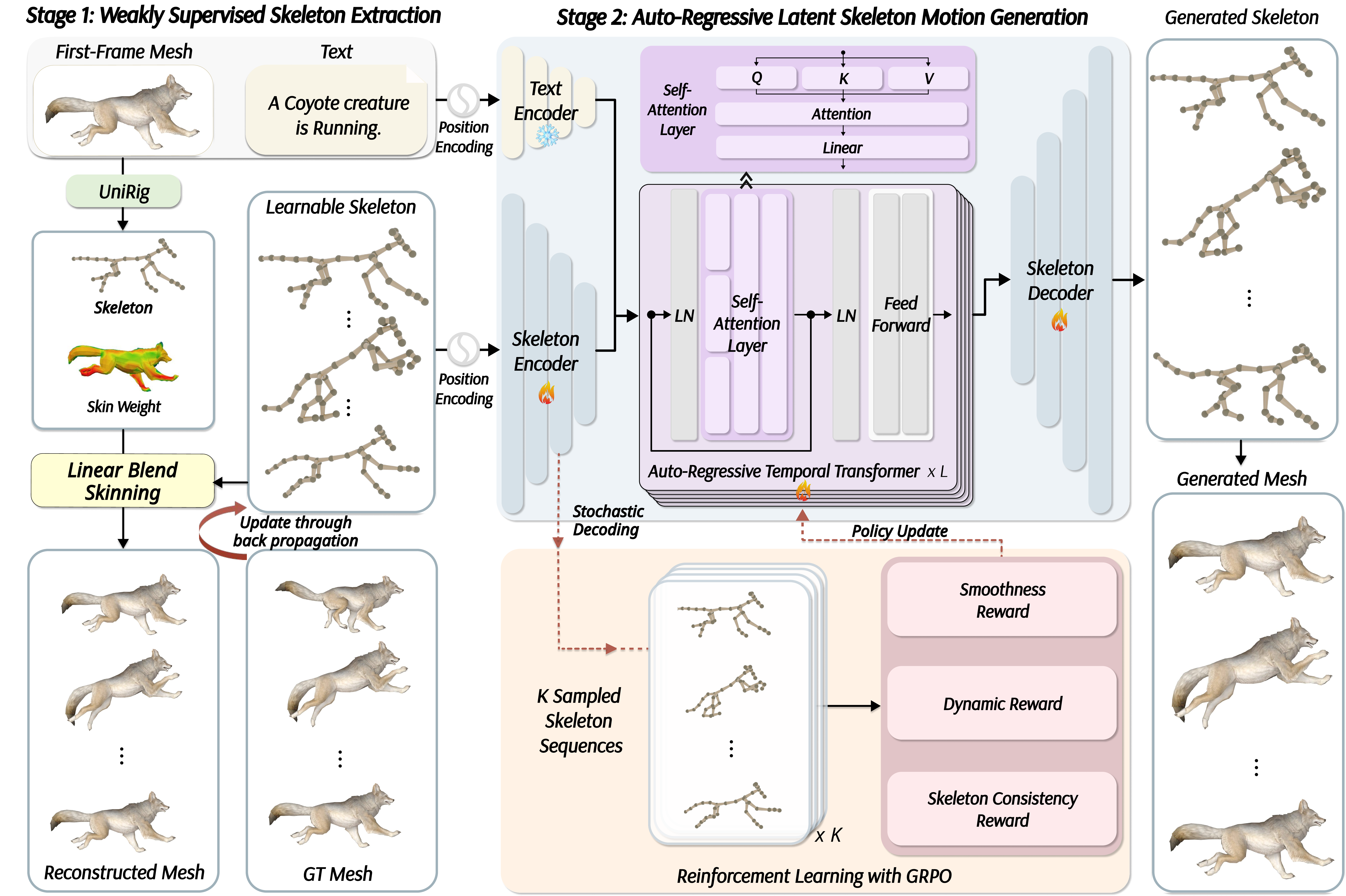}
\caption{
Overview of \textbf{SkelGen4D}.
Our framework consists of two stages.
\textbf{Stage 1: Weakly supervised skeleton extraction.}
Given the first-frame mesh, we apply automatic rigging to obtain the \jh{skeleton topology} and skinning weights $W$. Then we optimize per-frame skeleton $P_i$ via gradient-based fitting so that the mesh reconstructed by linear blend skinning $\hat{M}$ matches the ground truth mesh $M$ sequence, yielding temporally consistent pseudo-skeleton trajectories without per-frame skeleton annotations.
\textbf{Stage 2: Auto-regressive latent skeleton motion generation.}
Skeleton sequences $P$ are encoded into a latent space and generated autoregressively by a transformer conditioned on text.
To improve sequence-level motion quality under weak supervision, we introduce Motion-GRPO, which optimizes animation-aware rewards including temporal smoothness, bone-length consistency, and motion diversity.
At inference time, the generated skeleton motion is applied to the mesh via skinning to produce stable and editable 4D mesh animations.
}
     \label{fig:overview}
\end{figure*}

\subsection{Overview}

Given a text prompt $c$ and a \jh{first-frame} mesh $M_1$, our goal is to generate a temporally coherent 4D animation sequence $\{\hat{M}_t\}_{t=1}^T$ that follows the described motion while remaining editable and controllable through an explicit skeleton.
To achieve this, we adopt a two-stage training framework SkelGen4D, to generate articulated motion from weakly supervised mesh data.

\paragraph{Stage 1: Weakly-supervised skeleton extraction.}
As shown in Figure~\ref{fig:overview} ``Stage 1'', given a raw mesh animation sequence $\{M_t\}_{t=1}^T$, we extract temporally aligned pseudo-skeleton pose supervision $\{P_t\}_{t=1}^T$ via weakly supervised fitting.  
Starting from the first-frame mesh $M_1$, an initial skeleton pose $P_1$ and skinning weights $W$ are obtained through automatic rigging~\cite{zhang2025unirig}.  
We then optimize per-frame joint transformations $\{T_t\}_{t=1}^T$ such that
\[
\hat{M}_t = \mathrm{LBS}(M_1, P_1, W, T_t)
\]
matches the \jh{ground-truth} mesh $M_t$, converting raw mesh dynamics into temporally consistent pseudo-skeleton trajectories.

\paragraph{Stage 2: Auto-regressive latent skeleton motion generation.}
As shown in Figure~\ref{fig:overview} ``Stage 2'', using the extracted skeleton sequences, we train a text-conditioned auto-regressive model
$p(\{P_t\}_{t=2}^T \mid P_1, c)$ that generates full skeleton motion sequences conditioned on a first-frame pose $P_1$ and text prompt $c$.

\paragraph{Inference.}
At test time, given a new rest-pose mesh $M_1$ and a text prompt $c$, we first obtain its initial skeleton $P_1$ and skinning weights $W$ via automatic rigging.  
The trained motion model then generates a skeleton sequence $\{P_t\}_{t=2}^T$, which is applied to the mesh via linear blend skinning \jh{using the skinning weights $W$} to produce the animated mesh sequence $\{\hat{M}_t\}_{t=1}^T$.

\subsection{Stage-1: Weakly Supervised Skeleton Extraction}

\paragraph{Training Data and Objective.}
During training, our input is a raw dynamic mesh sequence
$\mathcal{M}=\{M_1,M_2,\ldots,M_T\}$, where $M_t$ denotes the mesh at time step $t$.
The goal of this stage is to recover temporally consistent skeleton pose sequences
$\mathcal{P}=\{P_1,P_2,\ldots,P_T\}$ that capture the observed mesh deformations and provide structured motion supervision for subsequent learning.
We assume that all frames in a sequence share a common skeletal topology and a
fixed vertex-to-joint skinning relationship, as articulated motion should be
driven by the same control structure throughout the animation.

\paragraph{Static Rigging Initialization.}
We treat the first frame $M_1$ as the canonical reference mesh.
Instead of estimating skeletons independently for each frame, which may lead
to topological inconsistency over time, we perform automatic rigging only once
on $M_1$ to obtain an initial skeleton pose $P_1$ and skinning weights:
\begin{align}
    P_1, \mathcal{W} = \mathrm{UniRig}(M_1).
\end{align}
The \jh{skeleton topology} implied by $P_1$ are \jh{fixed and shared across all frames}, while $\mathcal{W} \in
\mathbb{R}^{N \times J}$ denotes the vertex-to-joint skinning weights produced
by the rigging algorithm.
The \jh{skeleton topology} and skinning weights $\mathcal{W}$ are kept fixed for all frames and are not
optimized in this stage.

\paragraph{Dynamic Skeleton Pose Optimization.}
With the \jh{skeleton topology} and skinning weights $\mathcal{W}$ fixed,
we optimize per-frame skeleton poses $\{P_t\}_{t=2}^T$ to track the mesh motion over time.
At each time step $t$, the mesh is reconstructed from the canonical mesh $M_1$
using Linear Blend Skinning (LBS):
\begin{align}
    \hat{M}_t = \mathrm{LBS}(M_1, P_t, \mathcal{W}),
\end{align}
where $P_t$ parameterizes the joint transformations at frame $t$ given the \jh{skeleton topology}.

Concretely, LBS computes each deformed vertex as a weighted sum of joint transformations applied to its rest-pose position. We therefore recover the skeleton pose $P_t$ by fitting these joint transformations such that the
skinned mesh $\hat{M}_t$ best matches the \jh{ground-truth} mesh $M_t$.
To ensure temporal consistency, we initialize $P_t$ using the optimized pose from
the previous frame $P_{t-1}$ and then solve for the optimal pose by minimising a
reconstruction loss:
\begin{align}
    P_t^* = \arg\min_{P_t} \;
    \mathcal{L}_{\text{mesh}}(\hat{M}_t, M_t),
\end{align}
where $\mathcal{L}_{\text{mesh}}$ is defined as the mean squared vertex-wise
distance, \jh{and $P_t^*$ is the optimized pseudo-skeleton.}
\begin{align}
    \mathcal{L}_{\text{mesh}}(\hat{M}_t, M_t)
    = \frac{1}{N} \sum_{i=1}^{N}
    \left\| \hat{v}_i^{(t)} - v_i^{(t)} \right\|_2^2,
\end{align}
with $v_i^{(t)}$ and $\hat{v}_i^{(t)}$ denoting the $i$-th vertex of the ground-truth
and reconstructed meshes, respectively.
The pose parameters are optimized using gradient-based methods, with gradients
backpropagated through the LBS operation.

By iteratively optimizing this objective over the entire sequence, we convert
dense mesh deformations $\mathcal{M}$ into a temporally aligned skeleton pose sequence
$\mathcal{P}$, which serves as pseudo-ground-truth supervision for the subsequent
text-driven motion generation stage.

\subsection{Stage-2: Skeleton-based Auto-Regressive Generation}

Given the first-frame skeleton pose $P_1$ and a text prompt $c$, the goal of Stage~2
is to generate a temporally coherent skeleton pose sequence
$\mathcal{P}=\{P_1,P_2,\ldots,P_T\}$ in an auto-regressive manner.
At each time step, the model predicts the next pose conditioned on the previously
generated poses and the text condition, enabling controllable and variable-length
motion generation.
To handle varying skeleton topologies and joint counts across samples, we encode
each pose frame into a fixed-dimensional latent representation before the sequence
modeling.

\subsubsection{Skeleton Auto-Encoder}

We train a frame-wise skeleton auto-encoder
to map each pose frame to a compact latent code and reconstruct it back.
This provides a fixed-dimensional representation for sequence modeling while
supporting varying joint counts and heterogeneous skeleton topologies.

\paragraph{Pose Representation.}
A skeleton pose at time $t$ is represented as
$P_t \in \mathbb{R}^{J \times D_{\text{in}}}$, where $J$ is the number of joints and
$D_{\text{in}}=6$ denotes the joint parameterization (i.e., rotation and translation).
For batching, we apply zero-padding together with a validity mask to handle
variable joint counts.

\paragraph{Skeleton Encoder.}
We treat a pose frame as a set of joints and apply a shared point-wise MLP to encode each joint independently.
Then we adopt sinusoidal positional encodings to inject joint identity and structural cues.
The joint embeddings are aggregated using masked average pooling and projected
to a latent vector $z_t \in \mathbb{R}^{D}$:
\begin{align}
z_t = \mathrm{Proj}\!\left(
\frac{1}{\sum_{j=1}^{J} m_j}
\sum_{j=1}^{J} m_j \cdot
\mathrm{MLP}\!\left(P_t^{(j)} + \mathrm{pos}^{(j)}\right)
\right),
\end{align}
where $m_j \in \{0,1\}$ indicates whether joint $j$ is valid (i.e., non-padded), \jh{and $\mathrm{pos}^{(j)}$ denotes the positional encoded representation.}

\paragraph{Skeleton Decoder.}
A lightweight decoder maps the latent code back to joint features, 
producing a reconstructed pose $\hat{P}_t \in \mathbb{R}^{J \times D_{\text{in}}}$:
\begin{align}
\hat{P}_t = \mathrm{Dec}(z_t; J, m),
\end{align}
where the decoder outputs joint-wise predictions and ignores padded entries according to the mask.

\paragraph{Reconstruction Objective.}
The auto-encoder is trained to reconstruct the \jh{input pose frame $P_t$} by minimizing
a masked reconstruction loss \jh{$\mathcal{L}_{\text{ae}} (\cdot)$}:
\begin{align}
\mathcal{L}_{\text{ae}}(P_t,\hat{P}_t) =
\frac{1}{\sum_{j=1}^{J} m_j}
\sum_{j=1}^{J} m_j \cdot
\left\|\hat{P}_t^{(j)} - P_t^{(j)}\right\|_2^2.
\end{align}
The resulting latent code $z_t$ compresses skeleton pose semantics into a
fixed-dimensional representation, which is subsequently used by the temporal
transformer for auto-regressive motion generation.

\subsubsection{Latent Skeleton Motion Generation.}
As illustrated in Figure~\ref{fig:overview} ``Stage 2'', skeleton motion generation is formulated as an auto-regressive sequence prediction problem in the latent space.
Given frame-wise latent pose codes $\{z_t\}_{t=1}^{T}$ produced by the skeleton auto-encoder, the temporal encoder models their temporal evolution and predicts
future latent representations conditioned on past motion and text input.

\paragraph{Text Conditioning.}
The input text description \jh{$c$} is encoded using a pre-trained CLIP
model~\cite{radford2021learning}, producing a text embedding \jh{$c_{\text{text}}$} that captures high-level motion semantics.
This embedding is projected to the transformer hidden dimension via a linear layer \jh{$\psi(\cdot)$}:
\begin{align}
    c_{\text{text}} = \psi(\mathrm{CLIP}(c)).
\end{align}
The resulting text token provides a global conditioning signal for motion
generation across all time steps.

\paragraph{Auto-Regressive Temporal Transformer.}
The auto-regressive temporal transformer takes as input a sequence of tokens composed of the text
conditioning token and the latent pose codes from previous frames:
\begin{align}
    Z = [c_{\text{text}}, z_1, z_2, \ldots, z_{T-1}].
\end{align}
The text token is placed at the beginning of the sequence to condition the entire
motion trajectory.
A triangular attention mask is applied to ensure that the prediction at time step
$t$ depends only on the text condition and the latent history
$\{z_1,\ldots,z_{t-1}\}$.
The transformer outputs a sequence of context-aware latent predictions
$\{\hat{z}_t\}_{t=1}^{T}$, which are subsequently decoded into skeleton poses by the skeleton decoder.

Each transformer layer follows a standard residual attention block design.
Given input tokens $z^{0} = Z$, each layer is defined as:
\begin{align}
    f^{l} &= \mathrm{SelfAttn}(\mathrm{LN}(z^{l-1})) + z^{l-1}, \\
    z^{l} &= \mathrm{FFN}(\mathrm{LN}(f^{l})) + f^{l},
\end{align}
where $\mathrm{SelfAttn}(\cdot)$, $\mathrm{LN}(\cdot)$, and $\mathrm{FFN}(\cdot)$ denote multi-head
self-attention, layer normalization, and feed-forward networks, respectively.

\paragraph{Training Objectives.}
Stage~2 is trained with joint supervision in both the latent space and the pose space.
Specifically, the temporal transformer is supervised to predict latent pose codes
by minimizing the mean squared error between the predicted latent
$\hat{z}_t$ and the target latent $z_t$, where $z_t$ is extracted using a
pretrained and frozen skeleton encoder from Stage~1:
\begin{equation}
\mathcal{L}_{\text{feat}} =
\frac{1}{T} \sum_{t=1}^{T}
\left\| \hat{z}_t - z_t \right\|_2^2 .
\end{equation}

To ensure geometric plausibility of the generated motions, we additionally apply
a reconstruction loss on the decoded skeleton poses. The loss is evaluated only
over valid joints using a joint mask to handle variable joint counts:
\begin{equation}
\mathcal{L}_{\text{recon}} =
\frac{1}{T} \sum_{t=1}^{T}
\frac{1}{\sum_j m_j}
\sum_{j=1}^{J} m_j
\left\| \hat{P}_t^{(j)} - P_t^{(j)} \right\|_2^2 .
\end{equation}

The final training \jh{loss} is a weighted combination of both terms:
\begin{equation}
\mathcal{L}_{\text{total}} =
\lambda_1 \mathcal{L}_{\text{feat}} +
\lambda_2 \mathcal{L}_{\text{recon}} .
\end{equation}

\subsubsection{Reinforcement Learning with GRPO}

Although the temporal transformer is trained with frame-level latent and pose
supervision, such objectives do not explicitly model sequence-level motion
quality. To improve temporal coherence and kinematic plausibility, we further
introduce a reinforcement learning stage based on Group Relative Policy
Optimization (GRPO)~\cite{shao2024deepseekmath}.

For each input condition, we sample a group of $K$ motion sequences
$\{\mathcal{P}_i\}_{i=1}^{K}$ from the current policy via stochastic
auto-regressive decoding, and assign each sequence a reward defined over the
entire temporal horizon. The reward consists of three terms: smoothness,
dynamics, and skeleton consistency.

Let $\mathbf{x}_{t,j} \in \mathbb{R}^{3}$ denote the 3D position of joint $j$ at frame $t$, where $t=1,\dots,T$ and $j=1,\dots,J$. 
We first define velocity and acceleration as
\begin{equation}
\mathbf{v}_{t,j} = \mathbf{x}_{t+1,j} - \mathbf{x}_{t,j}, \qquad
\mathbf{a}_{t,j} = \mathbf{v}_{t+1,j} - \mathbf{v}_{t,j}.
\end{equation}

\paragraph{Smoothness.}
To encourage temporally smooth motion, we penalize large accelerations:
\begin{equation}
r_{\mathrm{smooth}}
=
-\frac{1}{(T-2)J}
\sum_{t=1}^{T-2}\sum_{j=1}^{J}
\|\mathbf{a}_{t,j}\|_2.
\end{equation}

\paragraph{Dynamics.}
To avoid degenerate solutions where joints remain nearly static, we introduce a dynamics reward that encourages sufficient motion. 
We measure the motion magnitude at each timestep as the aggregated velocity norm across all joints:
\begin{equation}
s_t =
\left\|
\mathrm{vec}\!\left(\{\mathbf{v}_{t,j}\}_{j=1}^{J}\right)
\right\|_2,
\qquad
\tilde{r}_{\mathrm{dyn}}
=
\frac{1}{T-1}\sum_{t=1}^{T-1} s_t.
\end{equation}
We further apply a thresholded penalty when the motion magnitude falls below a minimum level $\delta$:
\begin{equation}
r_{\mathrm{dyn}}=
\begin{cases}
\tilde{r}_{\mathrm{dyn}}-\gamma, & \text{if } \tilde{r}_{\mathrm{dyn}} < \delta,\\
\tilde{r}_{\mathrm{dyn}}, & \text{otherwise}.
\end{cases}
\end{equation}

\paragraph{Skeleton consistency.}
To preserve articulated structure, we enforce temporal consistency of bone lengths. 
Let $\pi(j)$ denote the parent joint of $j$, and $\mathcal{B}$ the set of valid bones. 
Using the first frame as reference, we define
\begin{equation}
\ell^{\mathrm{ref}}_{j}
=
\|\mathbf{x}_{1,j}-\mathbf{x}_{1,\pi(j)}\|_2 + \epsilon,
\qquad
\ell_{t,j}
=
\|\mathbf{x}_{t,j}-\mathbf{x}_{t,\pi(j)}\|_2.
\end{equation}
We compute the relative deviation
\begin{equation}
e_{t,j}
=
\frac{
\left|\ell_{t,j}-\ell^{\mathrm{ref}}_{j}\right|
}{
\ell^{\mathrm{ref}}_{j}
},
\qquad
\hat{e}_{t,j} = \max(e_{t,j}-\tau,\,0),
\end{equation}
and penalize excessive deviations:
\begin{equation}
\tilde{r}_{\mathrm{bone}}
=
-\frac{1}{|\mathcal{B}|\,T}
\sum_{j\in\mathcal{B}}\sum_{t=1}^{T}
\hat{e}_{t,j}^{\,2},
\qquad
r_{\mathrm{bone}} = \mathrm{clip}\!\left(\tilde{r}_{\mathrm{bone}},\, r_{\min},\, 0\right).
\end{equation}

The final sequence-level reward is
\begin{equation}
r
=
\lambda_{\mathrm{smooth}}\, r_{\mathrm{smooth}}
+
\lambda_{\mathrm{dyn}}\, r_{\mathrm{dyn}}
+
\lambda_{\mathrm{bone}}\, r_{\mathrm{bone}},
\end{equation}
where $\lambda_{\mathrm{smooth}}$, $\lambda_{\mathrm{dyn}}$, and
$\lambda_{\mathrm{bone}}$ balance the three reward terms.

Given rewards $\{r_i\}_{i=1}^{K}$ for a sampled group, GRPO computes the
normalized advantage
\begin{equation}
A_i = \frac{r_i - \mu(r)}{\sigma(r) + \epsilon},
\end{equation}
where $\mu(r)$ and $\sigma(r)$ are the mean and standard deviation within the
group. The policy is optimized by
\begin{equation}
\mathcal{L}_{\mathrm{GRPO}}
=
-\frac{1}{K}\sum_{i=1}^{K}
A_i \log p_{\theta}(\mathcal{P}_i \mid \mathbf{c})
+
\beta\,
\mathrm{KL}\!\left(
p_{\theta}(\cdot \mid \mathbf{c})
\,\|\, 
p_{\mathrm{ref}}(\cdot \mid \mathbf{c})
\right),
\end{equation}
where $\mathbf{c}$ is the input condition and $p_{\mathrm{ref}}$ is a frozen
reference policy. The sequence likelihood is factorized auto-regressively as
\begin{equation}
\log p_{\theta}(\mathcal{P}_i \mid \mathbf{c})
=
\sum_{t=1}^{T}
\log p_{\theta}(\mathbf{z}_{i,t} \mid \mathbf{z}_{i,<t}, \mathbf{c}),
\end{equation}
where $\mathbf{z}_{i,t}$ denotes the generated motion token or latent code at
step $t$. By optimizing sequence-level relative rewards, GRPO encourages motion
predictions that are smoother, more dynamic, and more kinematically consistent.

\begin{figure}
    \centering
    \includegraphics[width=1\linewidth]{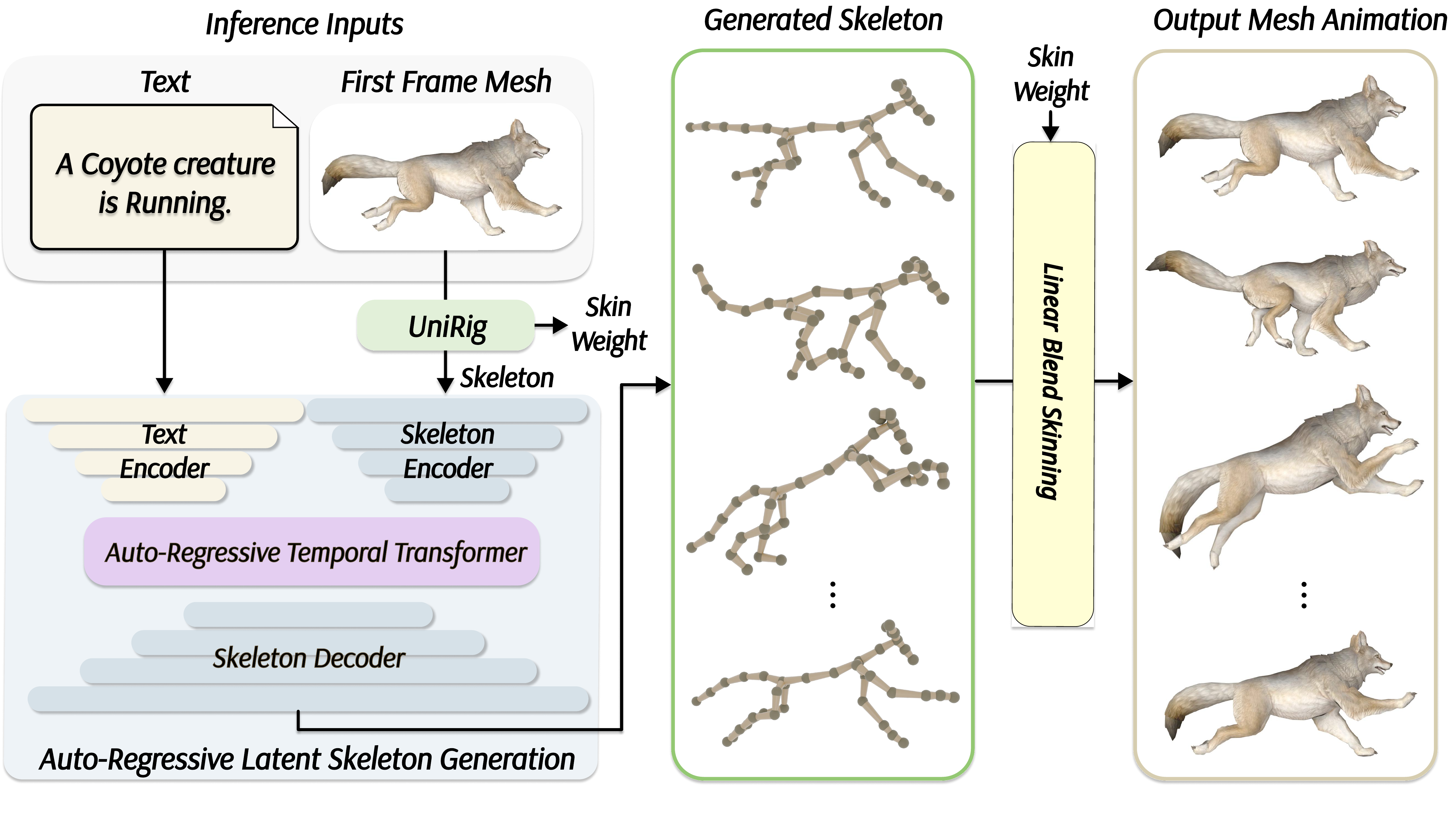}
     \caption{
Inference pipeline of \textbf{SkelGen4D}.
Given a text prompt and a first-frame mesh, we first obtain an initial skeleton and skinning weights via automatic rigging (UniRig).
The text and skeleton are then encoded and fed into an autoregressive transformer to generate a sequence of latent skeleton poses, which are decoded into explicit skeleton motions.
Finally, the generated skeleton sequence is applied to the mesh through linear blend skinning using the fixed skinning weights, producing temporally coherent and controllable mesh animations.
}
     \label{fig:inference}
\end{figure}

\subsection{Inference}

As shown in Figure~\ref{fig:inference}, at inference time, given a rest-pose mesh $M_0$ with a skeleton and skinning weights
$\mathcal{W}$ derived with UniRig~\cite{zhang2025unirig}, each predicted skeleton pose $\hat{P}_t$ provides per-joint local transformations to deform the mesh vertices via LBS:
\begin{equation}
\hat{v}_i^{(t)} = \sum_{j \in \mathcal{J}(i)} w_{ij} \, \mathbf{T}_j^{(t)} \, \bar{v}_i ,
\end{equation}
where $\bar{v}_i$ denotes the vertex position in the rest pose,
$w_{ij}$ is the skinning weight associating vertex $i$ with joint $j$, and
$\mathbf{T}_j^{(t)}$ denotes the global rigid transformation of joint $j$ at time $t$.

Applying this procedure independently at each time step yields a temporally
consistent animated mesh sequence
$\mathcal{M}^{o}=\{M'_1,\ldots,M'_T\}$.

\vspace{-3mm}
\section{Experiments}
\label{exp}
\subsection{Experimental Settings}

We evaluate our method on the Truebones Zoo dataset~\cite{Truebones} and the Diffusion4D dataset~\cite{liang2024diffusion4d}. Please refer to the Supplementary Material for details on the dataset, text prompt design, and evaluation metrics. 

\noindent\subsubsection{Datasets.}
We evaluate our method on the Truebones Zoo dataset~\cite{Truebones} and the Diffusion4D dataset~\cite{liang2024diffusion4d}.

Truebones Zoo contains a large collection of motion capture sequences with diverse articulated skeletons, spanning mammals, birds, insects, dinosaurs, fish, and snakes, with roughly 1,219 clips and 147,000+ frames across about 70 skeleton or mesh types.

Diffusion4D provides a large-scale, high-quality 4D dataset collected and curated from Objaverse-1.0 and Objaverse-XL, designed for spatial-temporal consistent 4D generation and supporting evaluation of dynamic mesh synthesis across varied object categories and motions. 

To handle varying sequence lengths and skeleton sizes, we adopt a padding strategy. Specifically, the number of joints is capped at 96, and skeletons with fewer joints are zero-padded accordingly. For temporal modeling, motion sequences within each batch are padded to the maximum sequence length in that batch.

\noindent\subsubsection{Text Prompt Design.}
For the Truebones Zoo dataset, we employ a structured text template to describe both the object category and its motion. Specifically, we prompt GPT to generate descriptions in the following form:
\emph{“This model represents a [object class] creature or object. This animation shows the [object class] performing the action [action name].”}

For the Diffusion4D dataset, where motion patterns are more diverse and less explicitly annotated, we adopt a concise visual description strategy. We instruct GPT with the following guidelines:
\emph{“You are a concise visual description assistant. Describe the visible motion or change between frames in one clear English sentence. Do not start with phrases like ‘The sequence shows’ or ‘It appears that’. Just describe the action directly.”}

\noindent\subsubsection{Evaluation metrics.}
For the Truebones Zoo dataset, following AnyTop~\cite{gat2025anytop}, we report four metrics that evaluate different aspects of the generated dynamic skeletons.
The metrics are computed independently for each skeleton, and the mean and standard deviation across all evaluated skeletons are reported as $mean^{\pm std}$:
(a) \emph{Coverage},
(b) \emph{Local Diversity},
(c) \emph{Inter Diversity}, and
(d) \emph{Intra Diversity Difference}.
Metrics (a) and (d) primarily assess fidelity to the ground-truth motions, while (b) and (c) evaluate motion diversity.
High fidelity with low diversity indicates overfitting, whereas low fidelity combined with high diversity suggests divergence and noisy generation.

For Diffusion4D, since different methods on it produce heterogeneous outputs (e.g., volumetric Gaussians, triangle meshes), it is difficult to apply unified geometric metrics.
To address this issue, we adopt two complementary evaluation protocols.
First, we follow recent practice and employ a vision-language model (VLM) as an automatic evaluator to assess animation quality at the perceptual and semantic level.
The VLM-based evaluation compares generated animations across methods in terms of \emph{geometric consistency}, \emph{structural consistency}, \emph{aesthetic quality}, and \emph{temporal coherence}, producing normalized scores that are comparable across different output formats. We distinguish between Dynamic Metrics (temporal behavior) and Static Metrics (geometric and visual quality at a specific moment).

\begin{table}
\caption{Quantitative results on the Truebones Zoo dataset. }
    \scalebox{0.9}{
    \centering
    \begin{tabular}{c|cccc}
        Model & Coverage$\uparrow$&Local Div.$\uparrow$ &Inter Div.$\uparrow$ &Intra Div. Diff.$\downarrow$\\
        \bottomrule
        MDM &$71.3^{\pm31}$ &$0.168^{\pm0.12}$ &$0.139^{\pm0.13}$ &$0.177^{\pm0.08}$\\
        SinMDM &$89.3^{\pm15}$ &$0.080^{\pm0.13}$&$0.280^{\pm0.13}$&$0.144^{\pm0.09}$\\
        Anytop &$80.5^{\pm20}$ &$0.252^{\pm0.14}$&$\textbf{0.312}^{\pm0.17}$&$0.118^{\pm0.07}$\\
        \bottomrule
Ours & $\textbf{90.52}^{\pm13}$ & $\textbf{0.253}^{\pm0.11}$& $0.271^{\pm0.17}$& $\textbf{0.088}^{\pm0.04}$\\
    \end{tabular}}
    \label{tab:truebone}
\end{table}

\begin{table}
\caption{Quantitative results (VLM-based) on the Diffusion4D dataset.}
    \scalebox{0.92}{
    \centering
    \begin{tabular}{lcccc}
            \toprule
            \textbf{Model} & \textbf{Geom.} $\uparrow$ & \textbf{Struc.} $\uparrow$ & \textbf{Aesth.} $\uparrow$ & \textbf{Temp.} $\uparrow$ \\
            \midrule
            Diffusion$^{2}$   & 6.34 & 6.08 & 5.12 & 4.52 \\
            Puppeteer         & 8.02 & 7.66 & 6.80 & 5.54 \\
            Animate AnyMesh   & 8.11 & 7.83 & 6.66 & 7.79 \\
            SS4D              & 8.23 & 8.42 & 6.93 & 8.11 \\
            MeshAction        & 8.13 & 8.11 & 7.00 & 8.23 \\
            \midrule
            \textbf{Ours}     & \textbf{8.62} & \textbf{8.52} & \textbf{7.55} & \textbf{8.39} \\
            \bottomrule
        \end{tabular}}
    \label{tab:diffusion4d_1}
\end{table}

\begin{table}
\caption{Human user study results on the Diffusion4D dataset.}
\vspace{-3mm}
    \scalebox{0.92}{
    \centering
     \begin{tabular}{lcccc}
            \toprule
            \textbf{Model} & \textbf{Geom.} $\uparrow$ & \textbf{Struc.} $\uparrow$ & \textbf{Aesth.} $\uparrow$ & \textbf{Temp.} $\uparrow$ \\
            \midrule
            Diffusion$^{2}$   & 4.83 & 4.50 & 3.50 & 4.67 \\
            Puppeteer         & 7.33 & 7.17 & 6.50 & 6.17 \\
            Animate AnyMesh   & 7.00 & 6.83 & 6.50 & 7.00 \\
            SS4D              & 6.00 & 5.67 & 5.00 & 5.83 \\
            MeshAction        & 7.33 & 7.26 & 7.17 & 6.83 \\
            \midrule
            \textbf{Ours}     & \textbf{7.67} & \textbf{7.50} & \textbf{7.67} & \textbf{7.50} \\
            \bottomrule
        \end{tabular}}
    \label{tab:diffusion4d_2}
\end{table}

\subsection{Analysis of Auto-Rigging and Skeleton Quality}

Since skeleton extraction is a core component of our pipeline, we provide a dedicated evaluation of its accuracy, robustness, and computational efficiency.

\paragraph{Success Rate}
We evaluate the robustness of our weakly supervised skeleton extraction by measuring the success rate over the dataset. 
A skeleton is considered successful if it satisfies two criteria: 
(i) it forms a valid articulated structure without topology collapse (e.g., no disconnected joints or invalid hierarchies), and 
(ii) it enables stable reconstruction of the input mesh sequence, measured by a reconstruction error below a predefined threshold. 
The success rate is computed as the percentage of sequences that satisfy both criteria. 
Our method achieves a success rate of approximately 85\% across diverse object categories, demonstrating strong robustness under weak supervision.

\paragraph{Discussion}
Although the skeleton extraction introduces a preprocessing cost, it enables the conversion of raw mesh animations into structured motion representations, which significantly improves the controllability and stability of downstream motion generation.
 
\subsection{Comparison with Existing Works on Truebones Zoo.}
\begin{figure*}
    \centering
    \includegraphics[width=0.9\linewidth]{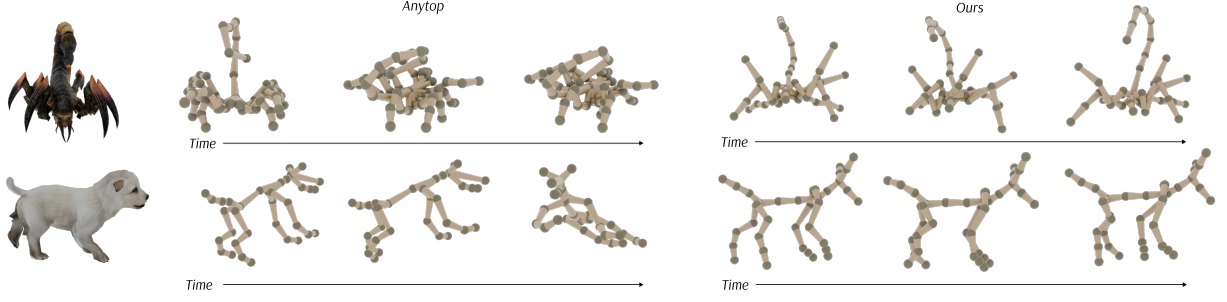}
     \caption{Qualitative comparisons of skeleton motion generation on the Truebones Zoo dataset.
}
     \label{fig:truebones}
\end{figure*}

\begin{figure*}
    \centering
    \includegraphics[width=0.95\linewidth]{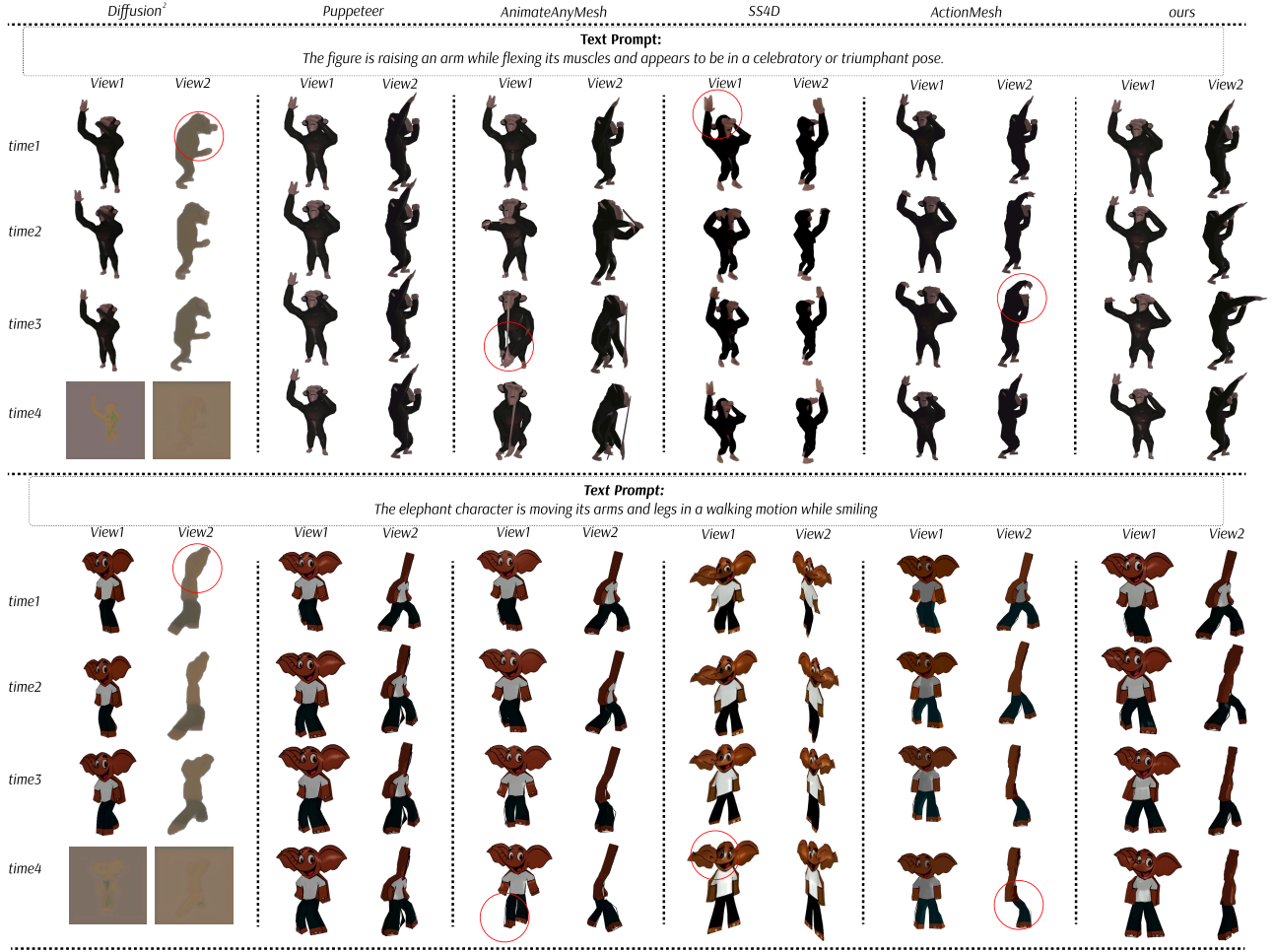}
     \caption{Qualitative comparisons of 4D generation on the Diffusion4D dataset. 
}
     \label{fig:diffusion4d}
\end{figure*}

\paragraph{Evaluation setting}
We first evaluate on the \textbf{Truebones Zoo} dataset, which provides high-quality ground-truth skeleton annotations across a wide range of articulated object categories. On this dataset, our evaluation focuses on the quality of the generated \emph{skeleton motions} instead of mesh appearance or rendering.
We compare our method against three representative skeleton-based motion generation models:
(i) \textit{MDM}~\cite{tevet2022mdm},
(ii) \textit{SinMDM}~\cite{raab2023single}, and
(iii) \textit{AnyTop}~\cite{gat2025anytop}.
For these baselines, we follow the evaluation protocol and report the results as provided in AnyTop~\cite{gat2025anytop}. 
Note that these \textbf{fully-supervised} methods rely on per-frame skeleton annotation, whereas our method is \textbf{weakly supervised} and does not utilize the ground-truth skeleton for training. Thus, our experimental setting is more challenging compared with the baseline approaches.

\paragraph{Quantitative comparison}
As shown in Table~\ref{tab:truebone}, even under weak supervision, our approach achieves state-of-the-art performance on Coverage and Intra Diversity Difference, indicating our model's superior performance in terms of motion fidelity. Meanwhile, our method attains comparable performance to existing fully-supervised approaches on Local Diversity and Inter Diversity, indicating the strong capability to generate diverse skeletal motions.

\paragraph{Qualitative comparison}
As shown in Figure~\ref{fig:truebones}, we qualitatively compare the generated skeleton sequences of our method with AnyTop~\cite{gat2025anytop}. Despite relying on frame-level skeleton annotations, AnyTop still exhibits noticeable structural instability and motion collapse in some cases. In contrast, our method produces stable and temporally consistent skeleton motions. These results suggest that our approach more reliably captures high-fidelity motion dynamics under weak supervision.

\begin{table}
    \caption{Ablation study on alternative weakly-supervised skeleton pseudo-label generation strategies.}
    \scalebox{0.9}{
    \centering
    \begin{tabular}{c|ccc}
         Mothed& Simulated Annealing & Least Squares& Gradient Descent\\
         \bottomrule
         MSE&  0.002&  0.12& $\textbf{2*10}^{\textbf{-5}}$\\
    \end{tabular}}
    \label{tab:ablation1}
\end{table}

\begin{table}
    \caption{Ablation study on Motion-GRPO.}
    \centering
    \scalebox{0.9}{
    \begin{tabular}{c|cccc}
         Model&Coverage$\uparrow$&Local Div.$\uparrow$ &Inter Div.$\uparrow$ &Intra Div. Diff.$\downarrow$\\
         \bottomrule
         w/o GRPO &$92.89^{\pm9}$&$0.215^{\pm0.09}$& $0.248^{\pm0.16}$&$0.090^{\pm0.07}$\\
 w GRPO&$\textbf{93.52}^{\pm13}$&$\textbf{0.260}^{\pm0.11}$& $\textbf{0.275}^{\pm0.17}$&$\textbf{0.085}^{\pm0.04}$\\
    \end{tabular}}
    \label{tab:ablation2}
\end{table}
\subsection{Comparison with Existing Works on Diffusion4D.}

\begin{figure*}
    \centering
    \includegraphics[width=1\linewidth]{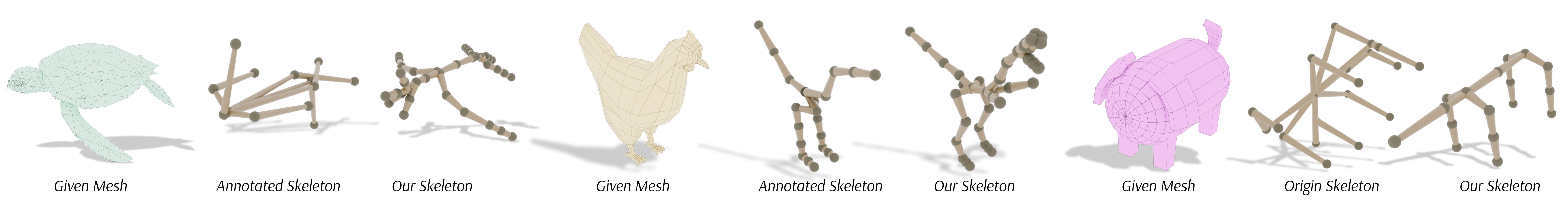}
     \caption{Comparisons of the annotated skeleton and the predicted pseudo skeleton using our SkelGen4D on the Diffusion4D dataset.
}
     \label{fig:pseudo}
\end{figure*}

\paragraph{Pseudo skeleton generation}
The Diffusion4D dataset provides large-scale mesh animation sequences but contains skeleton annotations that are often noisy, incomplete, or temporally inconsistent.
As illustrated in Figure~\ref{fig:pseudo}, the provided skeletons may exhibit incorrect topology, unstable joint placement, or \jh{insufficient} alignment
with the underlying mesh motion.
In contrast, our method derives temporally consistent pseudo-skeleton annotations directly from raw mesh sequences via weakly supervised fitting.
The resulting skeletons are structurally cleaner and better aligned with the observed mesh deformations than the original annotations.
These high-quality pseudo-skeletons subsequently serve as reliable supervision for second-stage motion modeling.


\paragraph{Evaluation setting}
As the skeleton annotations in Diffusion4D are noisy and unreliable, most recent approaches on this dataset adopt video-driven pipelines for animation generation. 
We compare our method against four representative baselines on Diffusion4D:
\textit{Diffusion$^2$} represents SDS-based volumetric 4D generation,
\textit{Puppeteer}, \textit{AnimateAnyMesh}, and \textit{SS4D} are most-recent video-driven mesh animation methods. 
For fair comparison, all baselines are evaluated using their official codes and pretrained models from GitHub.
Text-conditioned methods are directly driven by the same text prompts as ours.
For video-driven methods, we follow their standard pipelines and use rendered videos of the corresponding ground-truth 4D objects in Diffusion4D as driving inputs for 4D animation.

\paragraph{Quantitative comparison}
As shown in Tables~\ref{tab:diffusion4d_1} and~\ref{tab:diffusion4d_2}, our method consistently outperforms all baselines across \emph{geometric consistency}, \emph{structural consistency}, \emph{aesthetic quality}, and \emph{temporal coherence}, under both VLM-based evaluation and human user studies.
The consistent improvement across all criteria indicates that modeling motion through an explicit skeleton representation leads to more stable, coherent, and perceptually pleasing animations than SDS-based and video-driven approaches.
Overall, our method enables skeleton-driven animation on large-scale datasets that lack reliable skeleton annotations like Diffusion4D, significantly extending the usability of such datasets for controllable 4D generation.

\paragraph{Qualitative comparison}

Figure~\ref{fig:diffusion4d} presents qualitative comparisons on Diffusion4D.
Diffusion$^2$~\cite{yang2024diffusion2}, which relies on SDS-based volumetric optimization, produces severely blurred results in non-canonical views, indicating limited multi-view consistency and poor geometric fidelity.
Puppeteer~\cite{song2025puppeteer} exhibits very small motion amplitudes across frames, resulting in nearly static animations despite the input prompts. 
AnimateAnyMesh~\cite{wu2025animateanymesh} shows noticeable artifacts over time.
In the first example, abnormal connections appear between the hand and mouth regions, forming an unnatural vertical structure.
In the second example, the right leg fails to rotate coherently with the body, leading to a twisted and implausible pose.
%
SS4D~\cite{li2025ss4d} also suffers from structural degradation.
In the first example, the hand and face become fused, accompanied by severe blurring.
In the elephant example, the trunk and ears gradually collapse and merge in later frames, indicating accumulated deformation errors during generation.
ActionMesh~\cite{ActionMesh2026} also exhibits structural degradation during motion generation. 
In the first example, the generated motion shows significant arm distortion, where the arm bends unnaturally and fails to match the textual description. 
In the second example, the legs of the elephant character exhibit abnormal deformations during motion, deviating from the intended walking behavior described in the text.
In contrast, our method produces stable and coherent animations across all views and time steps.
By explicitly modeling motion in the skeleton space and applying skeleton-driven deformation, our approach preserves articulated structure, avoids deformation collapse, and maintains consistent motion patterns throughout the sequence. More results are provided in the file and video of the Supplementary Material.

\vspace{-1mm}

\subsection{Controllability}
\begin{figure*}[htbp]
  \centering
  \subfloat[]{
    \includegraphics[width=0.46\textwidth]{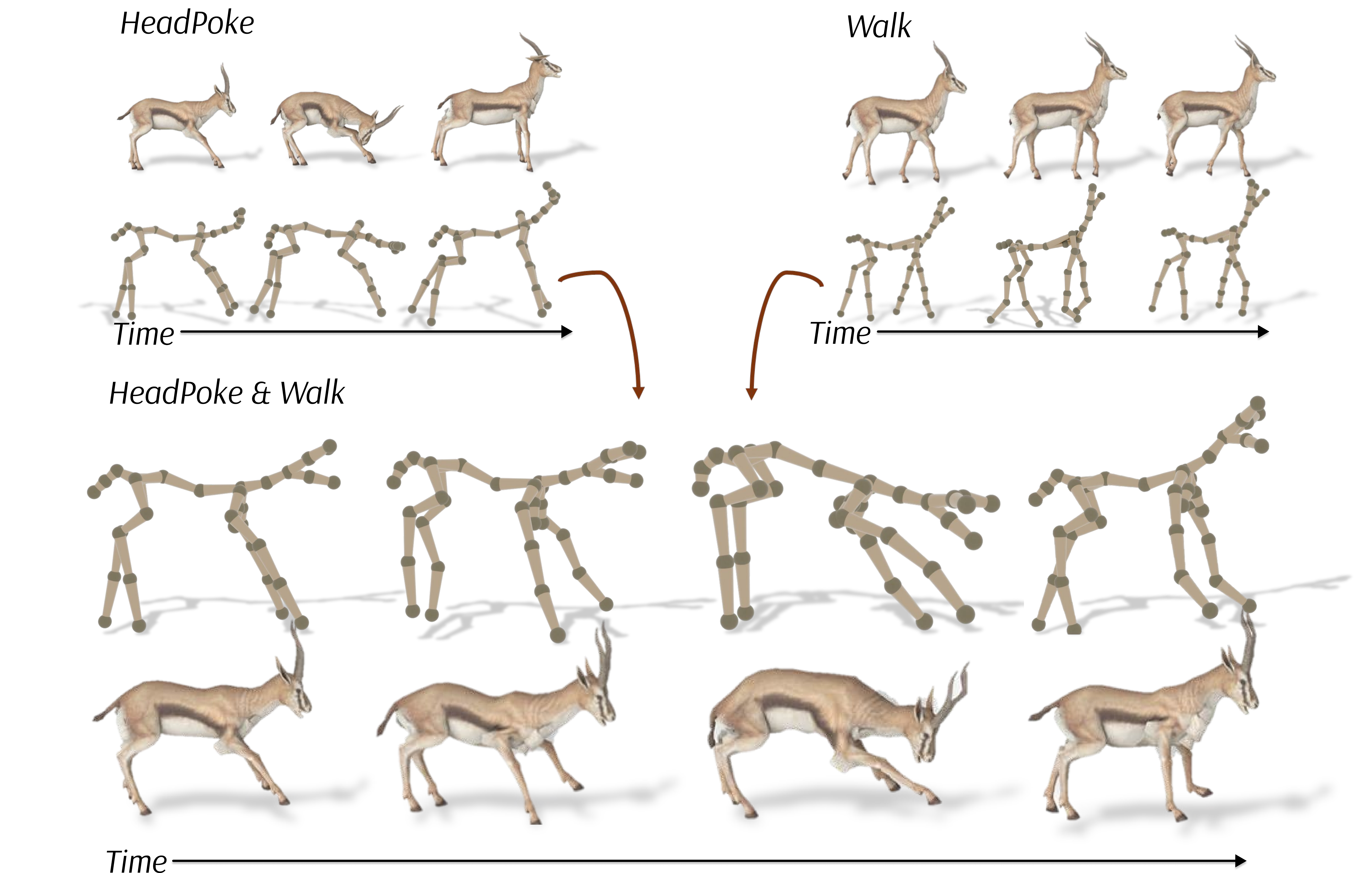}
  }
  \hfill
  \subfloat[]{
    \includegraphics[width=0.46\textwidth,height=5.3cm]{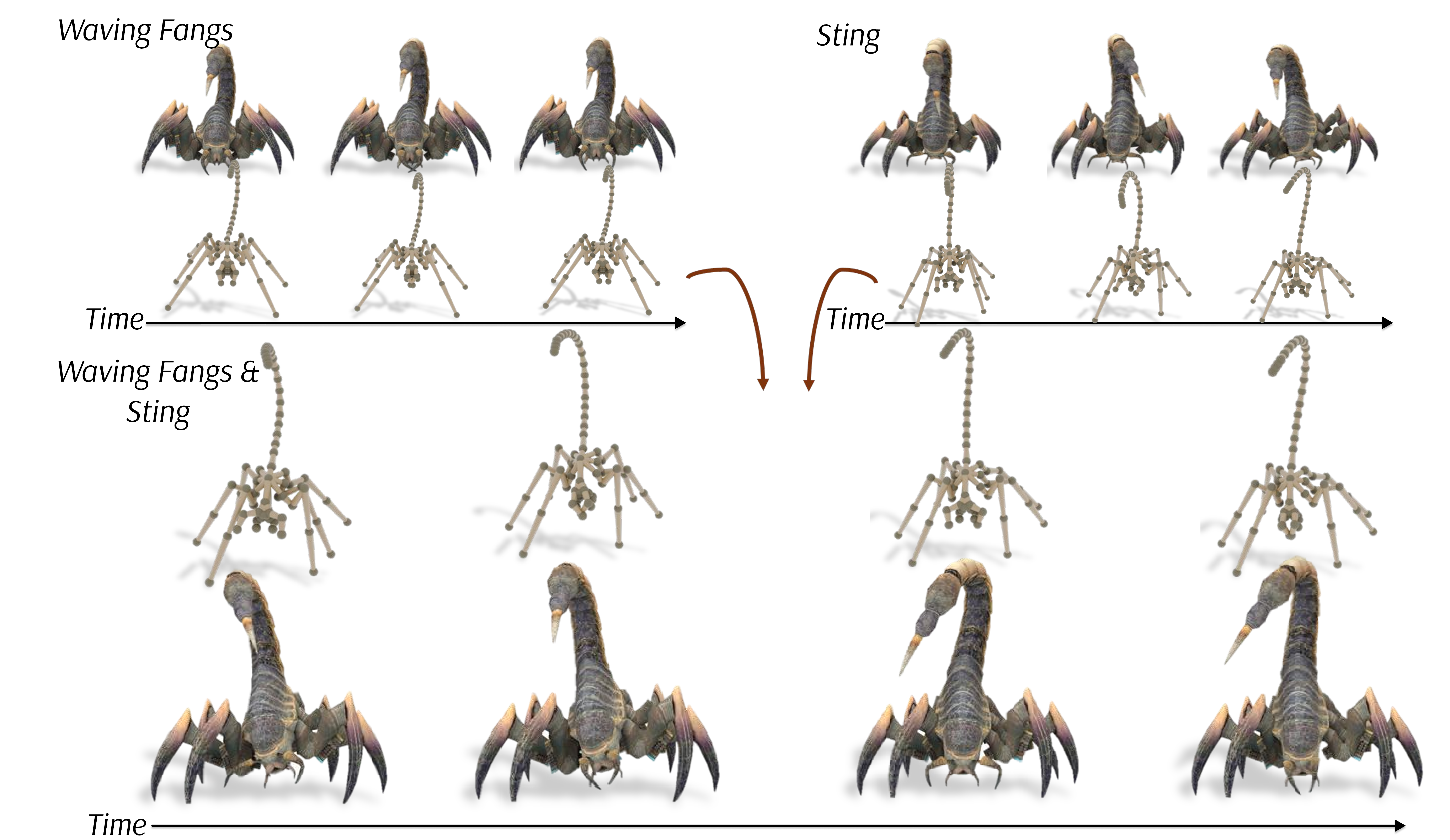}
  }

  \medskip
  \subfloat[]{
    \includegraphics[width=0.46\textwidth]{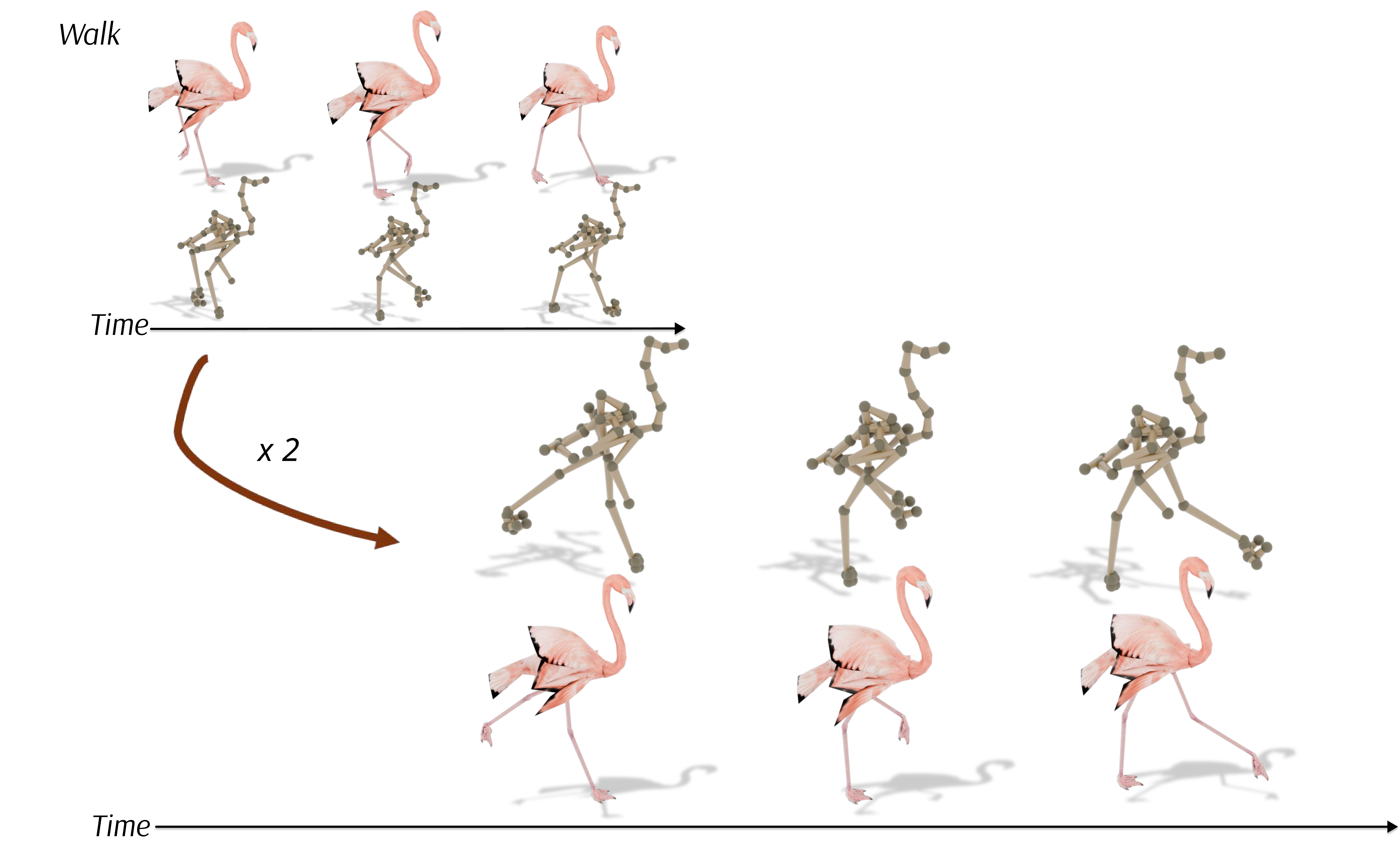}
  }
  \hfill
  \subfloat[]{
    \includegraphics[width=0.46\textwidth,height=5cm]{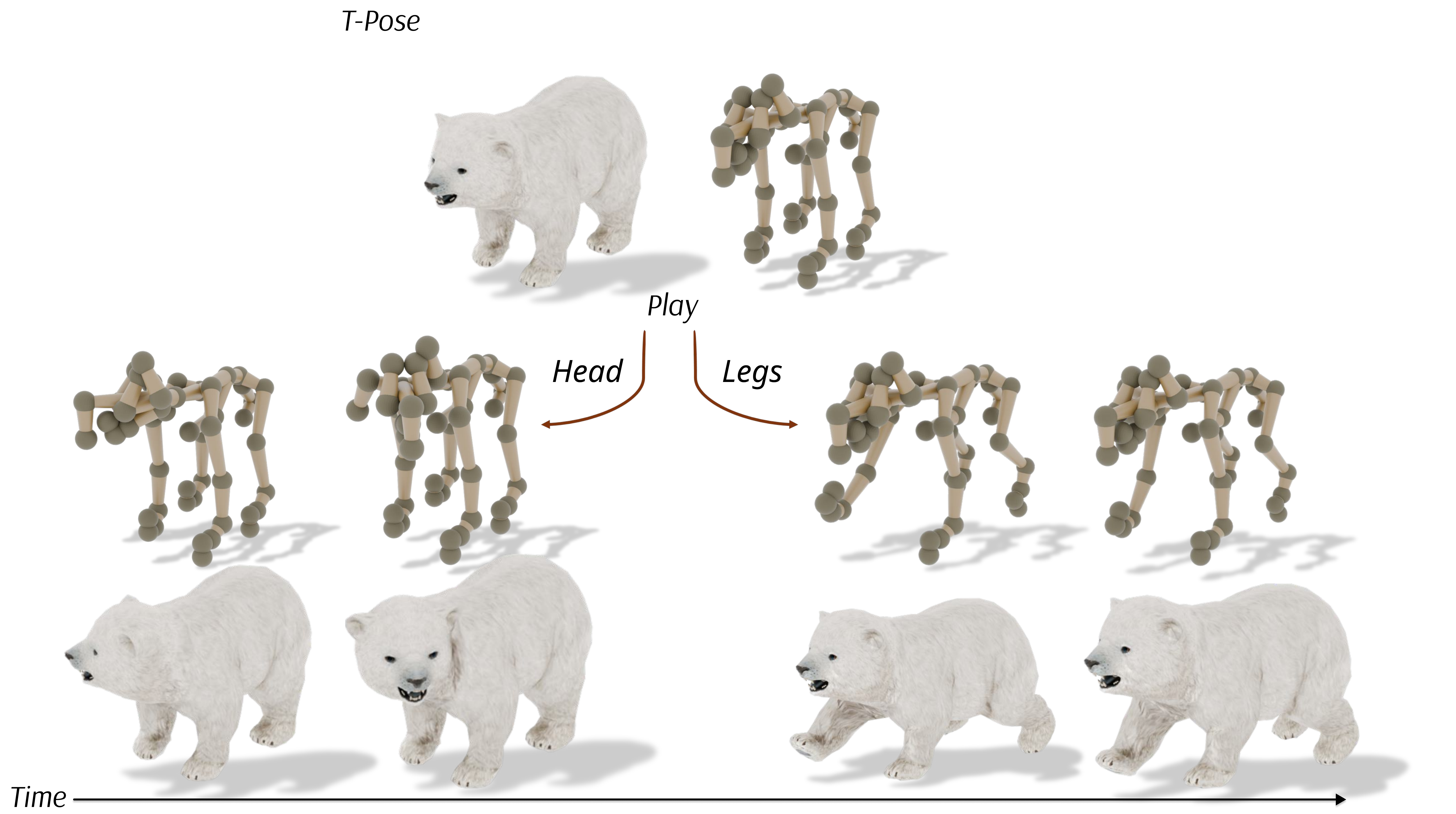}
  }

  \caption{Illustration of the controllability of our skeleton-based 4D generative framework SkelGen4D. (a,b) Motion composition. (c) Motion rescaling. (d) Local motion editing.}
  \label{fig:controllability}
\end{figure*}

This skeleton-level controllability aligns naturally with standard industrial animation workflows and supports flexible motion composition, rescaling, and editing. 
As illustrated in Figure~\ref{fig:controllability}, our skeleton-based framework enables flexible and fine-grained motion control that is difficult to achieve with video-driven methods.

\textbf{Motion composition.}
We first demonstrate motion composition by combining two independently generated motions, \emph{headpoke} and \emph{walk}, into a single animation.
The resulting skeleton cleanly separates behaviors across body parts: the upper body performs the head-poking motion while the lower body executes a walking gait, producing a coherent composite animation.

\textbf{Motion rescaling.}
We further show that the amplitude of skeletal motions can be smoothly rescaled in the skeleton space.
By adjusting joint rotations while preserving the underlying motion pattern, we generate animations with amplified or attenuated dynamics without introducing structural artifacts.

\textbf{Local motion editing.}
Finally, our representation supports localized motion editing.
As shown in the figure, we can modify limb motions while keeping the head static, or conversely edit head motion while freezing the limbs.
The part-specific control is enabled by explicit skeleton manipulation and is challenging for video-based methods, which typically entangle motion across the entire body.

\subsection{Ablation Studies}

\paragraph{Alternative Strategies for Weakly-Supervised Skeleton Generation}
To validate the effectiveness of our gradient-based formulation for generating pseudo-skeleton supervision, we design two alternative optimization strategies for skeleton-driven mesh fitting: \emph{simulated annealing} and \emph{least-squares optimization}.
All methods share the same initial skeleton and skinning weights, and aim to recover per-frame joint transformations that deform the canonical mesh to match the target mesh. Specifically, 
\textbf{simulated annealing} explores the pose space by stochastic sampling with a gradually decreasing temperature, while
\textbf{least-squares optimization} solves for the skeleton pose at each frame by minimizing the vertex reconstruction error using a closed-form or linearized least-squares solver.

\begin{figure*}
    \centering
    \includegraphics[width=1\linewidth]{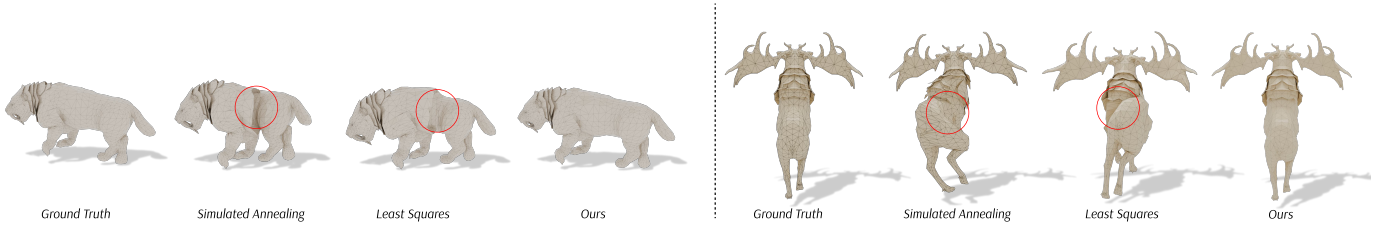}
     \caption{Qualitative ablation results comparing our SGD-based pseudo-skeleton generation with alternative strategies.
}
\label{fig:ablationnonstag1}
\end{figure*}

As shown in Table~\ref{tab:ablation1}, our gradient-based approach achieves orders-of-magnitude lower reconstruction error compared to alternative approaches. 
Figure~\ref{fig:ablationnonstag1} further provides qualitative comparisons, where alternative methods produce distorted skeletons, while our generated skeleton accurately recovers plausible motion.

\paragraph{Motion-GRPO}
To investigate the effectiveness of the proposed Motion-GRPO reinforcement learning strategy, we conduct an ablation study by removing the GRPO module and comparing the resulting performance with the full model. All other components and training settings are kept identical to ensure a fair comparison. As shown in Table~\ref{tab:ablation2}, removing Motion-GRPO leads to a noticeable degradation across multiple evaluation metrics. In contrast, Motion-GRPO introduces explicit reward-driven supervision on motion-related criteria, enabling the model to generate more diverse and dynamically coherent skeletal motions. 

\section{Conclusions, Limitations, and Future Works}

We have presented a skeleton-driven formulation for 4D generation that shifts the focus from implicit, instance-specific animation to explicit and structured motion modeling aligned with production pipelines.
Rather than synthesizing dynamics through volumetric optimization or video-conditioned deformation, our approach treats skeleton motion as a first-class representation that governs animation in a controllable and editable manner.
By recovering temporally consistent pseudo-skeletons from raw mesh animations, SkelGen4D enables learning articulated motion without relying on dense skeleton annotations, bridging large-scale data-driven generation with standard animation workflows.

Conceptually, this positions 4D generation closer to a structured motion design process, where articulated behaviors are modeled explicitly in skeleton space rather than being entangled with appearance or geometry-specific deformation.
Operating in this representation enables flexible motion control, composition, and editing, supporting a more modular and production-oriented paradigm for dynamic 3D content creation.
Through feed-forward skeleton motion generation, our framework demonstrates that high-quality, stable animations can be achieved under weak supervision while remaining compatible with industrial animation pipelines.

Meanwhile, the current formulation has several limitations.
Although skeleton supervision is weak, training still requires animated mesh sequences, and extending the framework to settings with only static geometry remains an open challenge.
In addition, the gradient-based skeleton fitting stage introduces a non-trivial preprocessing cost, and the quality of the extracted skeletons is bounded by the robustness of automatic rigging methods.
Finally, extremely complex or highly non-rigid motions may exceed the expressive capacity of the current skeleton-based formulation.
Addressing these challenges, particularly improving efficiency, robustness to rigging errors, and handling more complex motion patterns, offers promising directions for future work.

\bibliographystyle{IEEEtran}
\bibliography{reference}

\section{Biography Section}
\vspace{11pt}

\begin{IEEEbiography}[{\includegraphics[width=1in,height=1.25in,clip,keepaspectratio]{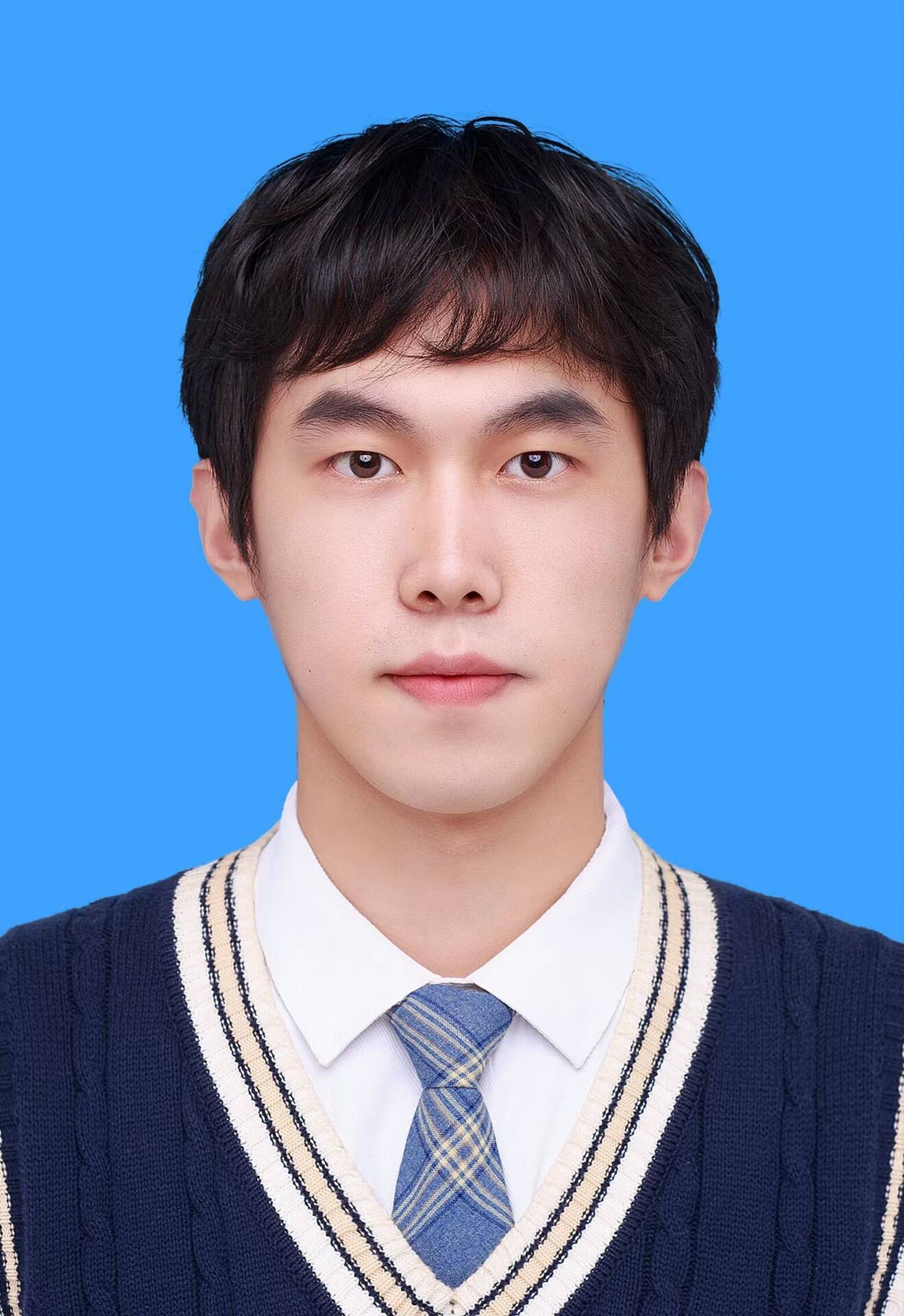}}]{Hao Feng}
Hao Feng received the B.S. degree from the Nanjing
University of Information Science and Technology
(NUIST), Nanjing, China, in 2022. He has entered
the school of computer science, Central China Normal University in 2023
to pursue a master’s degree. His research interests
include video comprehension and 4D reconstruction.
\end{IEEEbiography}
\vspace{-10mm}

\begin{IEEEbiography}                          [{\includegraphics[width=1in,height=1.25in,clip,keepaspectratio]{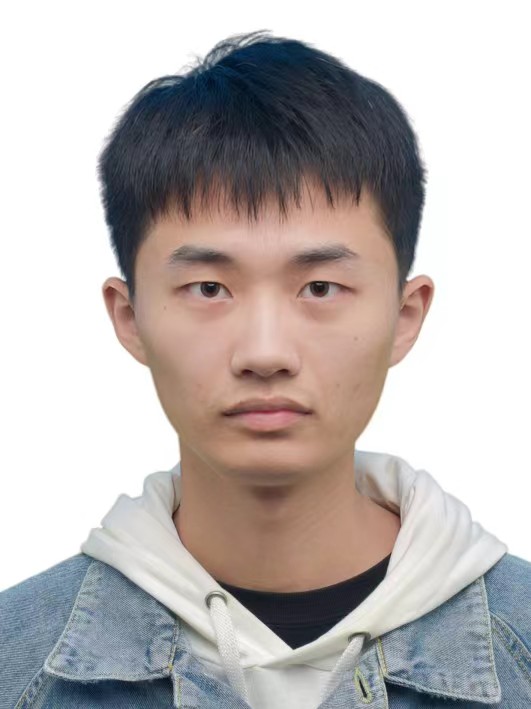}}]{Zhi Zuo} received the B.S. degree from the Nanjing University of Information Science and Technology (NUIST), Nanjing, China, in 2023. He has entered the College of Artificial Intelligence, Nanjing University of Aeronautics and Astronautics in 2023 to pursue a master's degree. His research interests include 3D point cloud analysis and artificial intelligence.
\end{IEEEbiography}
\vspace{-10mm}

\begin{IEEEbiography}                          [{\includegraphics[width=1in,height=1.25in,clip,keepaspectratio]{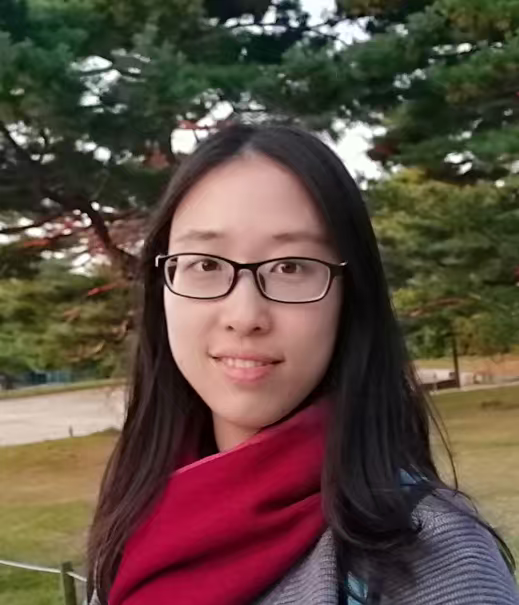}}]{Jia-Hui Pan} received the B.Sc. degree and the M.Phil degree at the School of Data and Computer Science from Sun Yat-Sen University. She is now a Ph.D candidate at the Computer Science and Engineering Department from The Chinese University of Hong Kong. Her research interests include robotics manipulation and 3D vision.
\end{IEEEbiography}

\begin{IEEEbiography}[{\includegraphics[width=1in,height=1.25in,clip,keepaspectratio]{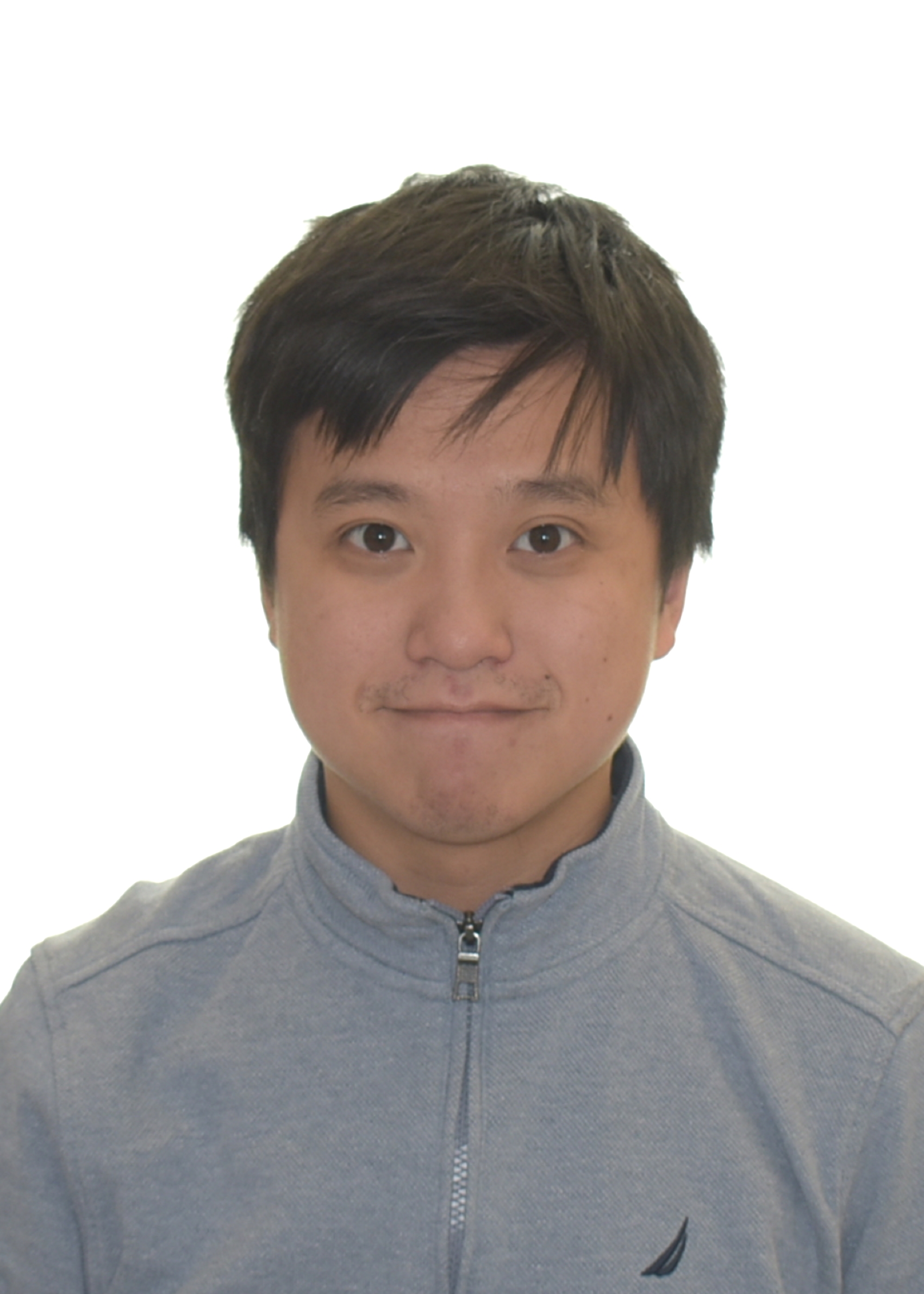}}]{Ka-Hei Hui} is a Senior AI Research Scientist at Autodesk AI Lab in Toronto. He received his Ph.D. from The Chinese University of Hong Kong, where he was supervised by Chi-Wing Fu. His research focuses on 3D shape analysis and generation, with an emphasis on developing deep learning methods for synthesizing 3D content in various representations, including point clouds, implicit functions, and meshes. He is also interested in data-driven approaches to combinatorial optimization problems, particularly in the context of computational assemblies.
\end{IEEEbiography}

\begin{IEEEbiography}[{\includegraphics[width=1in,height=1.25in,clip,keepaspectratio]{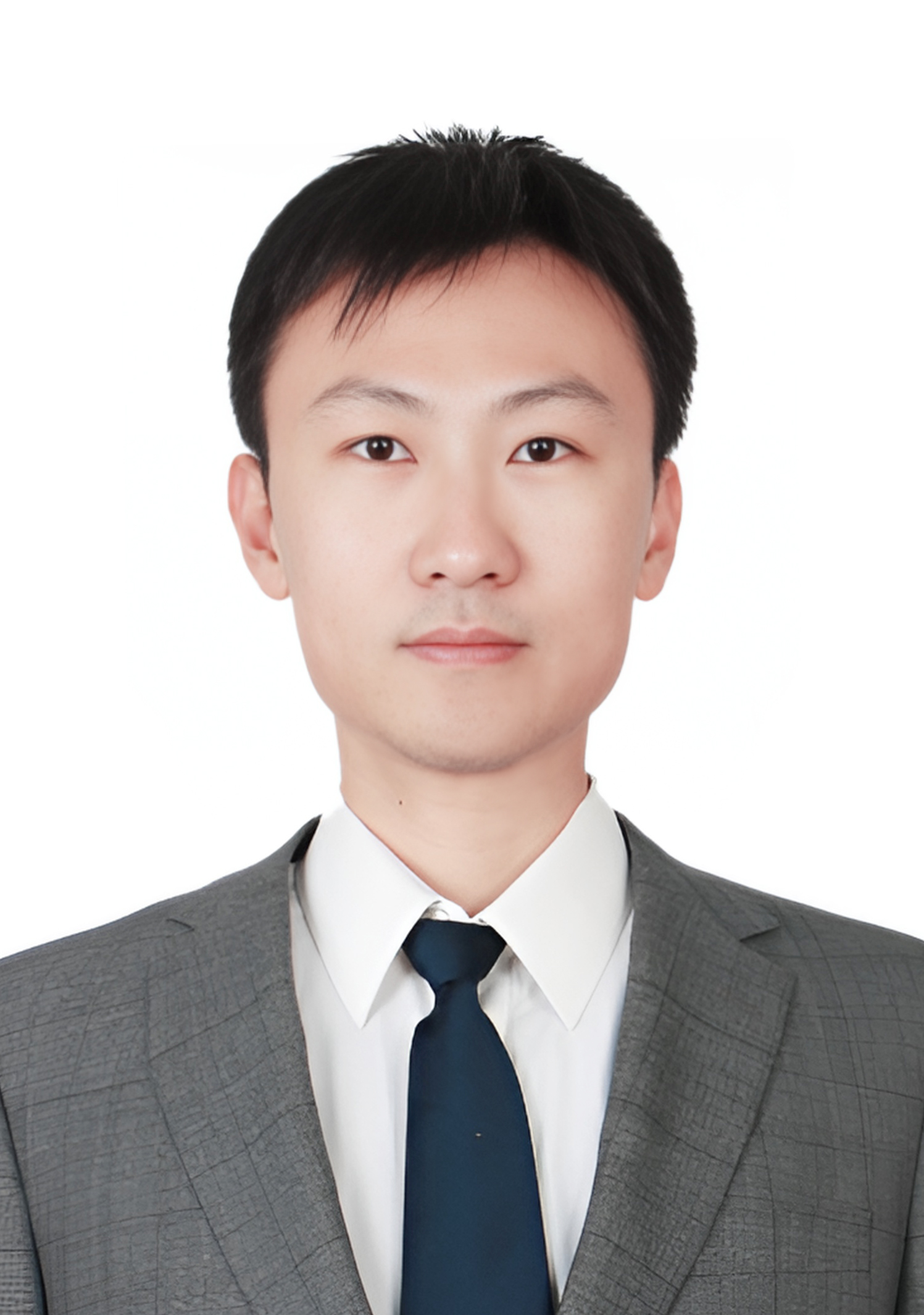}}]{Zhengzhe Liu} is currently an assistant professor at Lingnan University. He received his B.Eng degree in Information Engineering from Shanghai Jiao Tong University, and the M.Phil. and Ph.D. degree in Computer Science and Engineering from The Chinese University of Hong Kong. In 2024, he was a postdoctoral associate at Carnegie Mellon University. His research interests include AIGC, computer graphics, and 3D shape generation.
\end{IEEEbiography}

\begin{IEEEbiography}[{\includegraphics[width=1in,height=1.25in,clip,keepaspectratio]{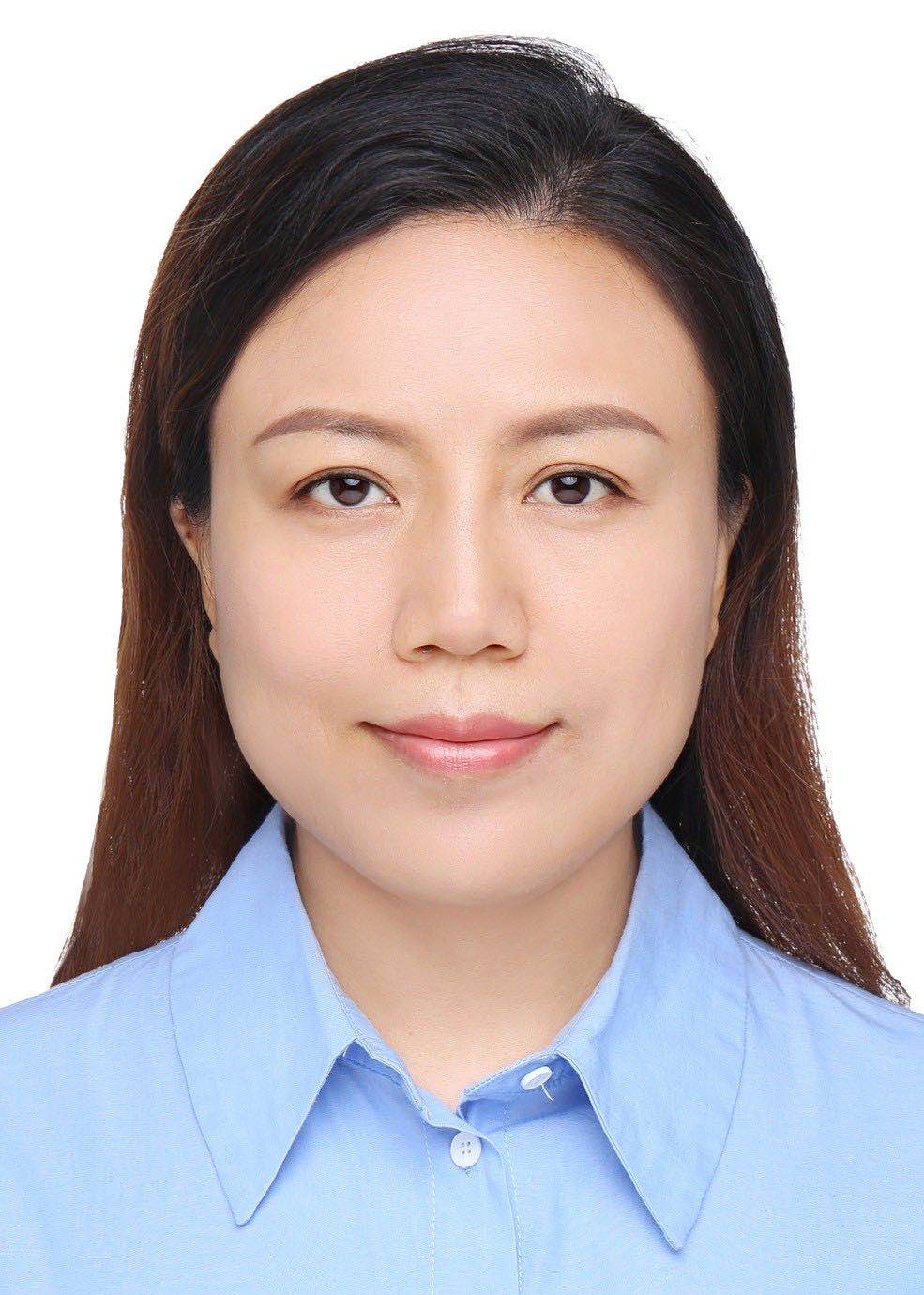}}]{ Dian Zhang} Prof. Dian Zhang is the vice dean and an associate professor of school of data science, Lingnan University. She is an honorary Professor of the University of Hong Kong. Before joining Lingnan, she was a Professor at the College of Computer Science and Software Engineering, Shenzhen University.  She was also an adjunct associate professor at Hong Kong University of Science and Technology (Guangzhou). She received her PhD in Department of Computer Science and Engineering, Hong Kong University of Science and Technology in 2010. Her research interests span over smart healthcare, big data analytics, mobile computing. She received 2022 Shenzhen 1st class Academic Research Outstanding Award in Nature Science, 2019 Ministry of Education of China 2nd Class Academic Research Outstanding Award in Nature Science, 2013 IBM Faculty Award, 2015 Shenzhen Peacock Talent Award and 2019 China Pacific Insurance Research Achievement Award.
\end{IEEEbiography}

\begin{IEEEbiography}[{\includegraphics[width=1in,height=1.25in,clip,keepaspectratio]{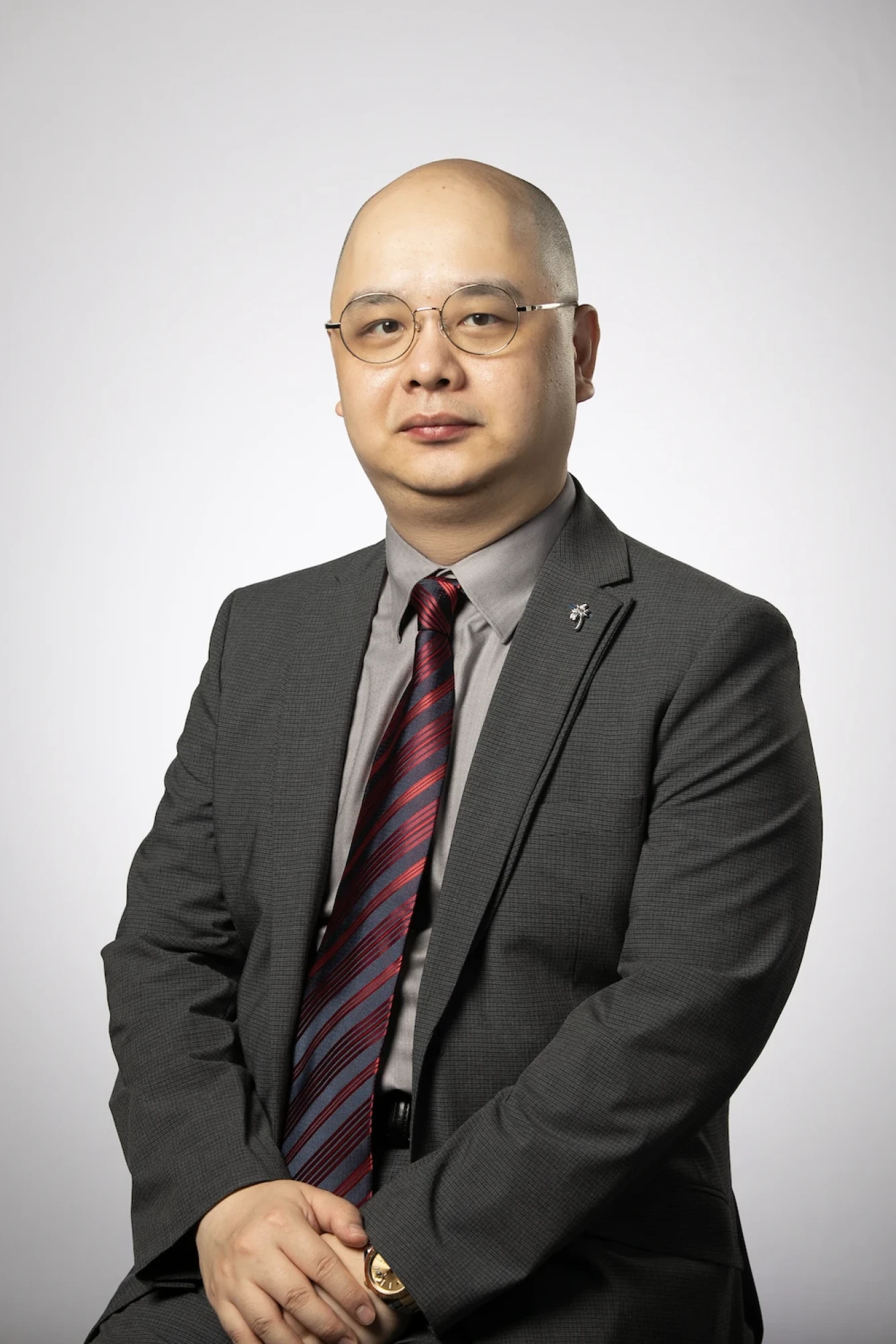}}]{Haoran Xie} (Senior Member, IEEE) received the Ph.D. degree in Computer
Science from City University of Hong Kong and an Ed.D degree in Language
Learning from the University of Bristol. He is currently a Professor and the
Person-in-Charge at the Division of Artificial Intelligence, Director of LEO Dr
David P. Chan Institute of Data Science, and Associate Dean of the School of
Data Science, Lingnan University, Hong Kong. His research interests include
natural language processing, large language models, and AI in education. He
has published 468 research publications, including 283 journal articles. He
is the Editor-in-Chief of Natural Language Processing Journal, Computers \&
Education: Artificial Intelligence, and Computers \& Education: X Reality. He
has been selected as the World’s Top 2\% Scientists by Stanford University.
\end{IEEEbiography}

\begin{IEEEbiography}
[{\includegraphics[width=1in,height=1.25in,clip,keepaspectratio]{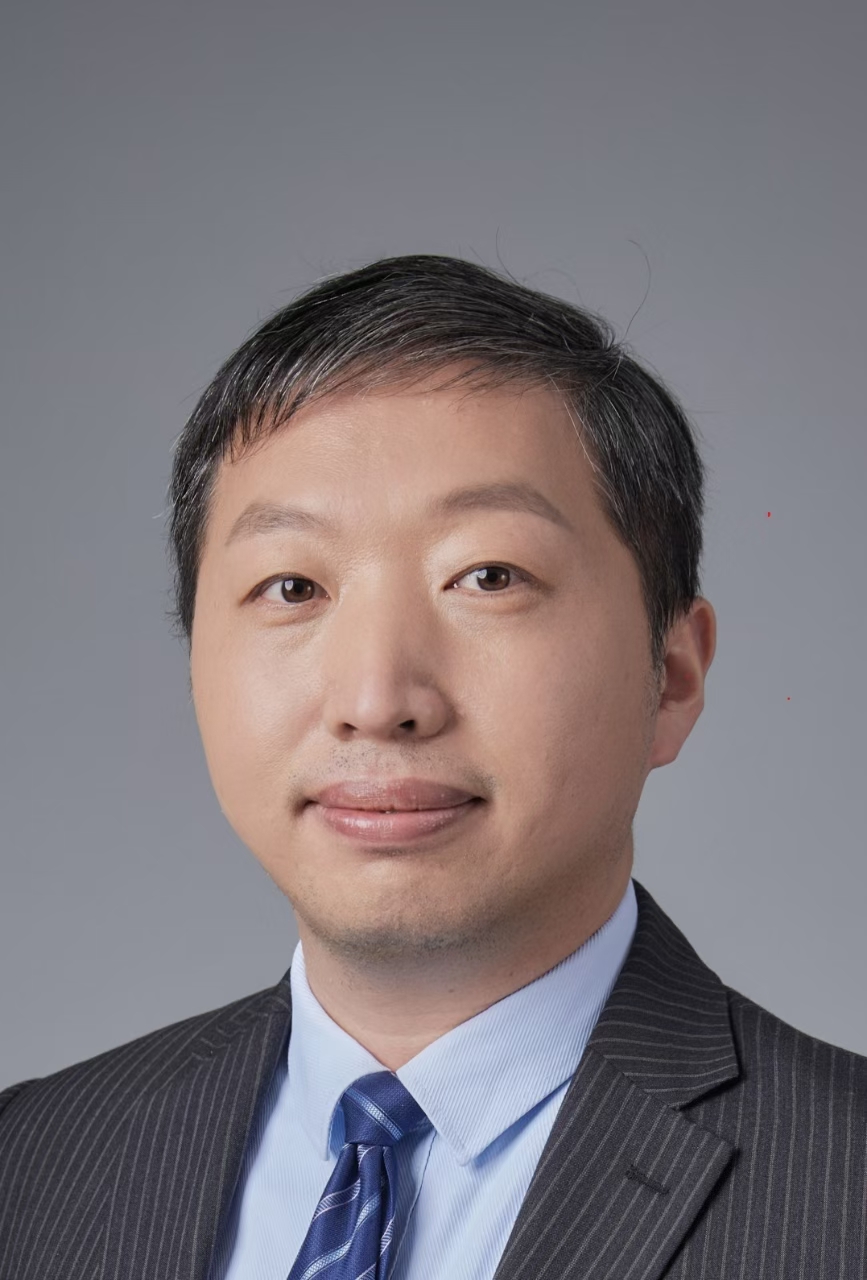}}]{Bin Sheng} obtained his Ph.D. degree in Computer Science and Engineering from The Chinese University of Hong Kong in 2011. He currently serves as a full Professor at the Department of Computer Science and Engineering, Shanghai Jiao Tong University. His research interests span virtual reality, machine learning, and medical data analysis. His work has been published in top-tier journals, including JAMA, Nature Medicine, Nature Biomedical Engineering, IEEE Transactions on Pattern Analysis and Machine Intelligence (TPAMI), IEEE Transactions on Visualization and Computer Graphics (TVCG), and International Journal of Computer Vision (IJCV).  Additionally, he was AI Challenge Co-Chair for DeepDRiD (ISBI 2020), DRAC (MICCAI 2022), and MMAC (MICCAI 2023). In recognition of his contributions, he received the Outstanding Contribution Award from the Computer Graphics Society in 2023.
\end{IEEEbiography}

\begin{IEEEbiography}
[{\includegraphics[width=1in,height=1.25in,clip,keepaspectratio]{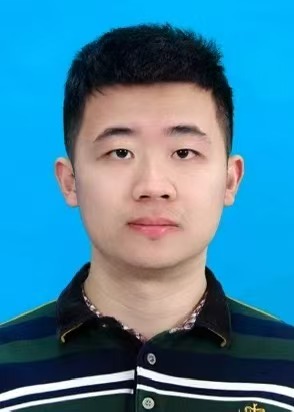}}]{Jingyu Hu} received his B.E. degree from the University of Chinese Academy of Sciences in 2020 and his Ph.D. degree from The Chinese University of Hong Kong in 2024. His research focuses on computer graphics and 3D vision.
\end{IEEEbiography}

\end{document}